\documentclass[wcp]{jmlr}

\usepackage{longtable}

\usepackage{booktabs}

\newcommand{\specialcell}[2][c]{%
  \begin{tabular}[#1]{@{}l@{}}#2\end{tabular}}
\makeatletter
\let\Ginclude@graphics\@org@Ginclude@graphics 
\makeatother

\title[Solving Machine Learning Problems]{Solving Machine Learning Problems}

\author{\Name{Sunny Tran} \Email{sunnyt@mit.edu} \addr MIT EECS\\
   \Name{Pranav Krishna} \Email{pkrishna@mit.edu}  \addr MIT EECS\\
   \Name{Ishan Pakuwal} \Email{ipakuwal@mit.edu} \addr MIT EECS\\
   \Name{Prabhakar Kafle} \Email{pkafle@mit.edu} \addr MIT EECS\\
   \Name{Nikhil Singh} \Email{nsingh1@mit.edu} \addr MIT Media Lab\\
   \Name{Jayson Lynch} \Email{jaysonl@mit.edu} \addr University of Waterloo\\
   \Name{Iddo Drori} \Email{idrori@mit.edu} \addr MIT EECS
  }

\begin{document}

\maketitle

\begin{abstract}
Can a machine learn Machine Learning? This work trains a machine learning model to solve machine learning problems from a University undergraduate level course. We generate a new training set of questions and answers consisting of course exercises, homework, and quiz questions from MIT's 6.036 Introduction to Machine Learning course and train a machine learning model to answer these questions. Our system demonstrates an overall accuracy of 96\% for open-response questions and 97\% for multiple-choice questions, compared with MIT students' average of 93\%, achieving grade A performance in the course, all in real-time. Questions cover all 12 topics taught in the course, excluding coding questions or questions with images. Topics include: (i) basic machine learning principles; (ii) perceptrons; (iii) feature extraction and selection; (iv) logistic regression; (v) regression; (vi) neural networks; (vii) advanced neural networks; (viii) convolutional neural networks; (ix) recurrent neural networks; (x) state machines and MDPs; (xi) reinforcement learning; and (xii) decision trees. Our system uses Transformer models within an encoder-decoder architecture with graph and tree representations. An important aspect of our approach is a data-augmentation scheme for generating new example problems. We also train a machine learning model to generate problem hints. Thus, our system automatically generates new questions across topics, answers both open-response questions and multiple-choice questions, classifies problems, and generates problem hints, pushing the envelope of AI for STEM education.
\end{abstract}

\section{Introduction} 
\label{section:introduction}
Significant progress has been made in natural language processing in question answering. The simplest questions to tackle, reading comprehension questions, are handled using large pre-trained Transformer models. More recent developments have allowed models to solve questions that require mathematical reasoning by predicting the associated equation. Unfortunately such models still fail on questions which involve basic Linear Algebra and Calculus, such as computing the length of a vector, which are prerequisites for Machine Learning. We create the first model that answers questions about Machine Learning, using Machine Learning.

\begin{table}
    \small
    \centering
    \begin{tabular}{|c|l|c|c|c|c}
    \hline
    \textbf{Week} & \textbf{Topic} & \textbf{Exp. Acc. ORQ} & \textbf{Acc. ORQ} & \textbf{Acc. MCQ} \\
    \hline
    \hline
        1 & Basics & 1.00 & 1.00 & 1.00\\
        2 & Perceptrons & 0.98 & 0.98 & 0.98\\
        3 & Features & 0.86 & 0.86 & 0.89\\
        4 & Logistic regression & 0.86 & 0.86 & 0.90\\
        5 & Regression & 0.97 & 0.97 & 0.97\\
        6 & Neural networks I & 1.00 & 1.00 & 1.00\\
        7 & Neural networks II & 0.97 & 0.97 & 0.98\\
        8 & Convolutional neural networks & 0.84 & 0.86 & 0.89\\
        9 & Recurrent neural networks & 1.00 & 1.00 & 1.00\\
        10 & State machines and MDPs & 0.94 & 1.00 & 1.00\\
        11 & Reinforcement learning & 1.00 & 1.00 & 1.00\\
        12 & Decision trees & 1.00 & 1.00 & 1.00\\
        \hline
        & Overall average over topics & 0.95 & 0.96 & 0.97\\
        \hline
    \end{tabular}
    \caption{Accuracy achieved using our system for each topic taught in MIT's Introduction to Machine Learning course, 6.036. Our system demonstrates an overall average expression accuracy (percent of correct expressions) of 95\% and value accuracy (percent of correct values) of 96\% for open response questions (ORQ), and accuracy (percent of correct values) of 97\% for multiple-choice questions (MCQ), achieving grade A performance in real-time.}
    \label{tab:results}
\vspace{-20pt}
\end{table}

\subsection{Solving Machine Learning Problems}
This work is the first to successfully solve Machine Learning problems (or questions) using Machine Learning. Specifically, our model handles the wide variety of topics covered in MIT's Introduction to Machine Learning course (6.036), except for coding questions and questions that require input images, which are shown in Table \ref{tab:results}, including basic linear algebra, perceptrons, features, logistic regression, regression, neural networks, convolutional neural networks, recurrent neural networks, reinforcement learning, and decision trees with overall grade A performance in real-time. Table \ref{results:examples}, along with the first 12 tables in Appendix A of the supplementary materials, show example input questions and output answers generated by our model for each machine learning topic. All of these questions are answered with an open (text) response. We also describe a method for taking advantage of multiple-choice questions, which increases overall accuracy from 94\% to 97\%.

While previous work has shown the ability of machine learning models to answer questions which require reading comprehension and mathematical reasoning, this work is the first to tackle machine learning problems using expression trees. A motivation for exploring learning machine learning is that STEM subjects have been shown to be more difficult than other topics for pure Transformer approaches \citep{hendrycks2020measuring}, and therefore require a rich representation. Recent work \citep{hendrycks2021measuring} demonstrates that current state-of-the-art Tranformers fail at few-shot learning these datasets, as well.

Beyond answering questions, our system is also capable of generating hints that help students learn. The ability to automatically generate hints benefits instructors, teaching assistants, and students, and suggests the ability for a machine learning model to aid learners within the subject.

\subsection{Related Work}

\subsubsection{Few-Shot Question Answering}
Recent work on Transformers and few-shot learning, such as GPT-3 \citep{brown2020language}, demonstrates the results of language models using few-shot learning for question-answering. Current work explores the effectiveness of such models in answering questions from a variety of academic subjects, ranging from the humanities and arts to science, technology, engineering, and math (STEM) subjects \citep{hendrycks2020measuring}. In this dataset, one of the most difficult topics for Transformers are STEM with topics such as Formal Logic, Abstract Algebra, and High-school Math barely scoring above random chance. Our work goes beyond pure Transformers, to generate question answers using expression trees.

\subsubsection{Math Word Problems}
A math word problem is defined as a collection of sentences that form a question for which there is a single correct numerical answer. The goal of our work is to construct a model that correctly answers the question. The solution to a math word problem is obtained by applying mathematical operations to numerical values available in the problem, numerical values derived from them, or constants that are associated with the problem text, such as $\pi$ or $e$. Math word problems are primarily problems of comprehension, which motivates the use of Transformers. For humans, the performance of solving math word problems is connected with working memory capacity and inhibitory skills. Humans may generate a graph representation for word problems as a heuristic to reduce the problem complexity \citep{verschaffel2020word}. The ability to select among different graph representations and the knowledge about such different representations is referred to as meta-representational competence \citep{verschaffel2020word}. Solving a math word problem may require associating numerical values in the text with entities, using the relationship between multiple entities and numerical values to find a solution. Solutions to math word problems are often in the form of an expression tree. In addition, math word problems have also been included in datasets which are used to benchmark the performances of machine learning models \citep{hendrycks2020measuring, hendrycks2021measuring}.

Given a math word problem, early work \citep{kushman2014learning} learns to select a template equation using a hand-designed set of features and aligns the numerical values to the template equation. Recent work \citep{xie2019goal,wu2020knowledge,qin2020semantically} directly generates the expression tree representing an expression which answers the math word problem. Developing the expression tree decoding method further, other work \citep{zhang2020graph} introduces a graph representing the relationships between words and quantities and a graph representing the relationships between different quantities. Graph-to-tree neural networks \citep{li2020graph} consist of a graph encoder and hierarchical tree decoder, mapping structured inputs to outputs. Recent work on expression pointer Transformers \citep{kim2020point} generate expression tokens consisting of operators and operands to overcome the fragmentation of expressions and the separation between operands and context. Expression pointer Transformers \citep{kim2020point} use an efficient ALBERT Transformer with state-of-the-art results. The methods used by these works follow a general approach: encoding the input question, processing the encoded information, decoding the processed encoding, and evaluating the decoded information. Such an approach demonstrates the ability of mathematical reasoning. Our work for solving machine learning problems uses an expression tree representation within an encoder-decoder framework. In contrast with all previous work on solving math word problems, our approach naturally handles questions involving recursion.

\subsubsection{Graph Neural Networks}
Graph neural networks (GNNs) \citep{kipf2017semi,hamilton2017inductive,velivckovic2018gat,xu2019powerful} are commonly used for predicting properties of particular nodes, edges, sub-graphs or the entire graph. In this work, we use GNNs for predicting an entire graph property, which is the encoding of a tree representing an expression.

\subsubsection{Transformers}
The development of the Transformer is perhaps the most significant recent breakthrough in natural language processing \citep{devlin2019bert,shoeybi2019megatron,raffel2020exploring,brown2020language} and computer vision \citep{dosovitskiy2020image}. Transformers are built from attention mechanisms \citep{vaswani2017attention}. An unsupervised corpus of text is transformed into a supervised dataset by defining content-target pairs along the entire text. Transformers are trained with the objectives of predicting masked target words in sentences and predicting whether two sentences follow each other. A language model is first trained to learn an embedding of words or sentences, followed by a map from low dimensional content to target \citep{mikolov2013efficient}. This embedding is then used on a new, unseen and small dataset in the same low-dimensional space. Our work incorporates Transformers for both augmenting machine learning questions and solving them. Each of our models consists of a Transformer for encoding text. We use the HuggingFace Transformer library \citep{wolf2020huggingface} and OpenAI GPT-3 \citep{brown2020language} for comparisons.

\subsubsection{Hints}
An example of previous work on hint generation uses a factory model \citep{hintfactory}. A Markov Decision Process (MDP) generator is used, along with a hint provider, to articulate hint in text format. The state space of hints is reduced using abstract syntax trees. A similar approach applies automatic hint generation for programming problems \citep{Riversetal} using large corpuses of previous students work, which provides information about most common correct solutions.

\section{Methods}

In this section, we describe our dataset, system architecture, and testing methodology. The key components leading to our success are: (i) a dataset of Machine Learning questions annotated with expression trees representing their solutions; (ii) a data augmentation technique which allows us to automatically generate new, related questions; and (iii) the use of Transformers and graph neural networks to generate expression tress for solving problems, rather than just predicting numerical answers.

\subsection{Dataset and Data Augmentation}

The questions used for training and testing our model are drawn from MIT's Introduction to Machine Learning course, also known as 6.036, which is available online\footnote{\href{https://introml.odl.mit.edu}{https://introml.odl.mit.edu}}.
Due to the novelty of training a machine learning model to answer Machine Learning questions, we curate a new dataset from 6.036 exercises, homeworks, and quizzes. Example questions and answers are shown in Table~\ref{results:examples}. Questions involving an image or coding questions are not used in our dataset, and are the limitation of our approach. We present example collected questions in the supplementary material.

\paragraph{Expression Trees} We further enhance this dataset by including not only answers to the question, but also expression trees representing how the answer is calculated from information in the question. We find that a relatively small set of expression trees is sufficient to cover most questions. Nodes in the tree are either quantities that appear in the question, special constants, or mathematical operations. The allowed set of operations are $\{ +, -, /, \times, \max, \log, \exp \}$. The supplementary material consist of a comprehensive set of expression trees our system automatically generates from machine learning questions.

\paragraph{Data Augmentation} After collecting questions from 6.036 homework assignments, exercises, and quizzes, each question is augmented by paraphrasing the question text. This results in five variants per question. Each variant, in turn, is then augmented by automatically plugging in 100 different sets of values for variables. Overall, we curate a dataset consisting of 14,000 augmented questions. Even though there are only 28 canonical expression tree structures, this represents a large space after augmentation and scales up. We provide an example in the supplementary material which shows an example of questions generated by augmentation: (i) question; (ii) template; (iii) paraphrased template; (iv) paraphrased question (1 out of 5 questions); (v) augmentation by plugging in examples values (1 of 100 questions) into each paraphrased template; (vi) final expression; and (vii) answer.

\subsection{Architecture}
\begin{figure*}
	\centering
	\includegraphics[width=\textwidth]{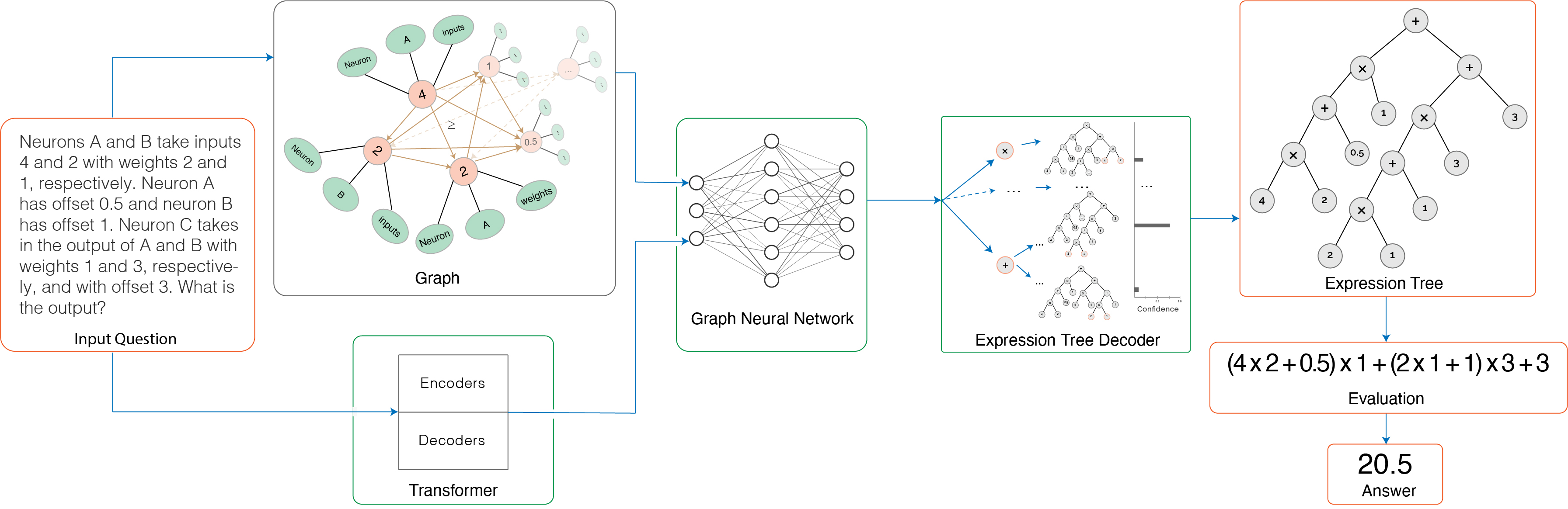}
	\caption{Architecture for solving Machine Learning questions. We create the question text embedding by a Transformer and also a graph representing the relationship between words and values and among values. These are passed through a graph neural network to create a vector in latent space which is then decoded to generate expression trees. The process of inputting the same question into the model is repeated a fixed number of times, resulting in multiple outcomes due to dropout in the GNN. The expression tree with the highest confidence is parsed to get the expression and the final value.}
	\label{fig:arch}
\end{figure*}

Figure \ref{fig:arch} illustrates our architecture. The model inputs a question which uses a math expression to solve it and returns the answer to the question. This is done by using modules within our architecture. Each input question is passed into a Transformer model to encode the question and also a graph which parses the question into its words and numerical components. The Transformer creates an embedding for the input question. The graph collects the words and numerical values of the question and formulates relationships between words and values and between values in the question. The embedding of the question from the Transformer and the graph are passed together into a GNN to create an embedding, which is subsequently decoded by a Tree Decoder to generate an expression tree. The model evaluates the generated expression tree and returns the computed value as output. 

\paragraph{Graph Representation}
A Machine Learning question consists of words and quantities, and our method represents it as an expression tree which upon evaluation provides an answer. In order to capture the relationship between the problem text and the output tree, we represent the question as a graph where the nodes represent values and words, and the edges represent relationships between nodes as shown in Figure \ref{fig:arch}. Our model finds the maximum-likelihood estimate tree given the constructed graph and the possible node values.

\paragraph{Expression Tree Decoder} A key feature of our architecture is the generation of an expression tree for each question which is then evaluated to produce an answer. The decoder generates the tree in a depth-first fashion using the current node and the latent encoding of the tree to generate either an operator or number for that node and to create children if applicable.

\subsection{Additional Components}
Our primary architecture is aimed at answering open-response numerical questions. Here we describe the alterations made to handle multiple choice questions and problem hint generation.

\subsubsection{Solution Ranking}
To answer multiple-choice questions, we sample our model multiple times to create a solution distribution, rank each solution in order of the frequency of appearance, and select the most frequent solution as the answer. Our architecture reflects this algorithm by performing a beam-search to return a list of answers from most to least confident. Although our model is able to produce results which are not dependent on answer choices, we find it important that our model take advantage of this extra information. This naturally lends itself to finding the most probable expression tree, conditioned on the fact that it must evaluate to one of the answers.

We also seek to create a system which generates incorrect answer questions close to the correct answer, similar to human-created multiple choice exams. We define an answer as “close” to the correct answer if a series of small perturbations of the correct expression tree yields another expression tree that evaluates to that answer. We consider two types of perturbations: rotating the tree, and adjusting the leaf nodes. Rotations are similar to those of balanced binary search trees, and are performed for trees with at least $5$ nodes. We also consider adjusting leaf nodes by multiplying or adding a constant value. Both strategies are used to create sufficient incorrect answers to present at the model.

\subsubsection{Hint Generation}
In addition to solving questions, we also generate hints for questions answers. Our generated hints provide a range of outputs, from classifying questions by topic, to partial solutions, to full solutions, as well as provide partial or full examples with different values for the variables. Both partial solutions and example solutions with other values help students fill-in missing knowledge gaps and learn by example. By enabling the model to generate hints, we are able to leverage the understanding of the model to aid the understanding of learners stumped by Machine Learning questions. This provides a more responsive and scalable system than relying on instructor assistance alone.

The generated hints in our model are categorized into three types:
\begin{itemize}
	\item Expression hints: include expressions or sub-expressions with variables from the question. Each hint is an expression which is generated using a partial solution of the question.
	\item Value hints: include expressions or sub-expressions with numerical values from the question. These are types of hints that directly provide information from the values used in the question.
	\item Example hints: include expressions or sub-expressions with values different from those found in the problem. These are generated using different values from the ones used in the question, demonstrating steps of the solution by examples.
\end{itemize}

\begin{figure}[h!]
	\centering
	\includegraphics[width=0.6\textwidth]{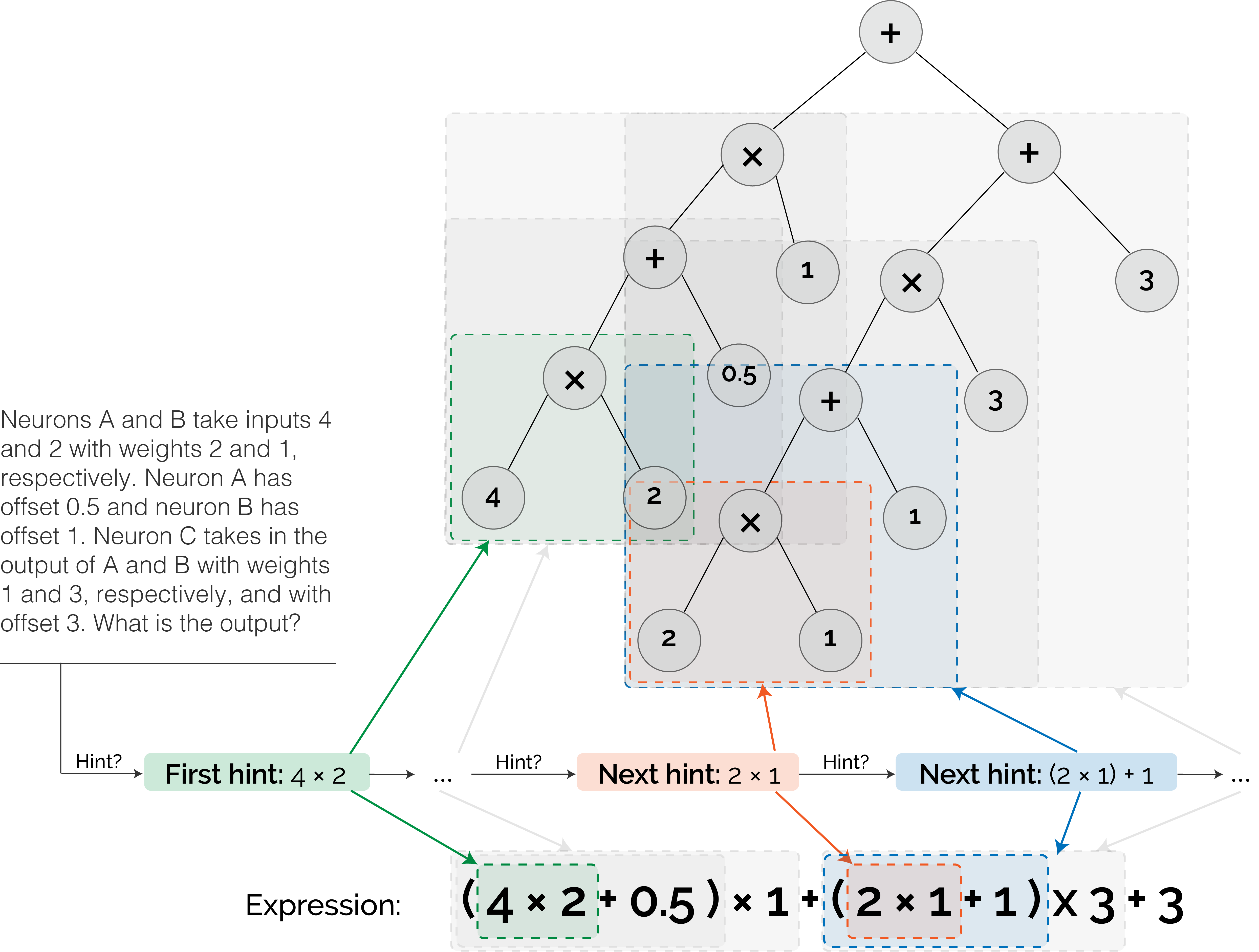}
	\caption{Generating successive hints for a machine learning problem. The expression tree is traversed in a binary search tree order starting from the left-most non-leaf node to give the sub-expression rooted at each non-leaf node as hint.}
	\label{fig:hint-nn}
\end{figure}

Figure \ref{fig:hint-nn} depicts self-contained hints and the progression of providing hints to the learner until a full solution is be achieved.

\subsection{Model Evaluation}
The evaluation of our model is two-fold; we determine its accuracy in reformulating the expected mathematical expression to solve the problem, as well as its accuracy in achieving the correct computed result. We refer to these performance metrics as:
\begin{itemize}
	\item Expression accuracy: number of answers with the correct expression.
	\item Value accuracy: number of answers with the correct numerical value.
\end{itemize}

We take random disjoint training and test sets with an 0.8/0.2 split. We determine the accuracy of our model by calculating the percentage of questions which the model gets correct. To determine the accuracy per topic, we collect a mapping between each topic and the templates for that topic. We recover the topic of a question which has been correctly responded to by removing numerical values and representations in the question and matching it to one of the original templates. 

\paragraph{Class Details and Grading}
MIT's 6.036 Introduction to Machine Learning \footnote{MIT 6.036 website: introml.odl.mit.edu.}, has around 500 students. Graded assignments for the course include:

\begin{itemize}
    \item Exercises: short quizzes submitted before lectures.
    \item Quizzes: timed questions that do not contain coding questions.
    \item Homeworks: mostly coding problems.
\end{itemize}

Quiz and homework questions are typically more difficult than exercise questions. We use questions from all the above, excluding questions which consist of input images that are required for the solution and excluding coding questions. Overall, around 58\% of the questions are multiple choice, and 42\% are open questions. Exercise questions do not consist of required input images or coding problems.

For exercises, MIT students are given an unlimited number of attempts at a problem. If they ever respond with the correct answer they receive $1$ point for the problem, otherwise they received a $0$. To prevent our model from simply brute-forcing answers our grading scheme is as follows:
\begin{itemize}
	\item For open-response questions (ORQ): we give our model one chance to provide an answer, and if the answer is correct, we get 1 point, otherwise no points. 
	\item For multiple-choice questions (MCQ): with $n$ choices and the correct answer chosen at attempt $t$, we give our model $\frac{n-t}{n-1}$ of a point.
\end{itemize}

\subsection{Implementation Details}
Our code is in Python, using the libraries Pytorch\footnote{https://github.com/pytorch/pytorch}, Deep Graph Library (DGL)\footnote{https://github.com/dmlc/dgl}, HuggingFace's Transformers\footnote{https://github.com/huggingface/transformers}, and comparisons using the OpenAI API\footnote{https://beta.openai.com}. DGL is primarily used for its graph convolutional network representation, while the Transformers library provides the necessary tokenizers and pre-trained models. All experiments are run on a standard cloud instance optimized for compute. Training time is around a day, augmentation time is less than a minute, whereas test time is milliseconds per question. We make our code available in the supplementary material.

Our implementation takes roughly 3 hours to train, including data augmentation, model training, and model evaluation. Data augmentation takes less than 30 seconds to perform, creating 14,000 questions from 140 paraphrased questions. Model training takes about 6.5 minutes per epoch, and evaluating the model on the 2,800 test examples takes about 1.5 minutes.

\section{Results}

\begin{figure}[h!]
    \centering
    \includegraphics[width=\textwidth]{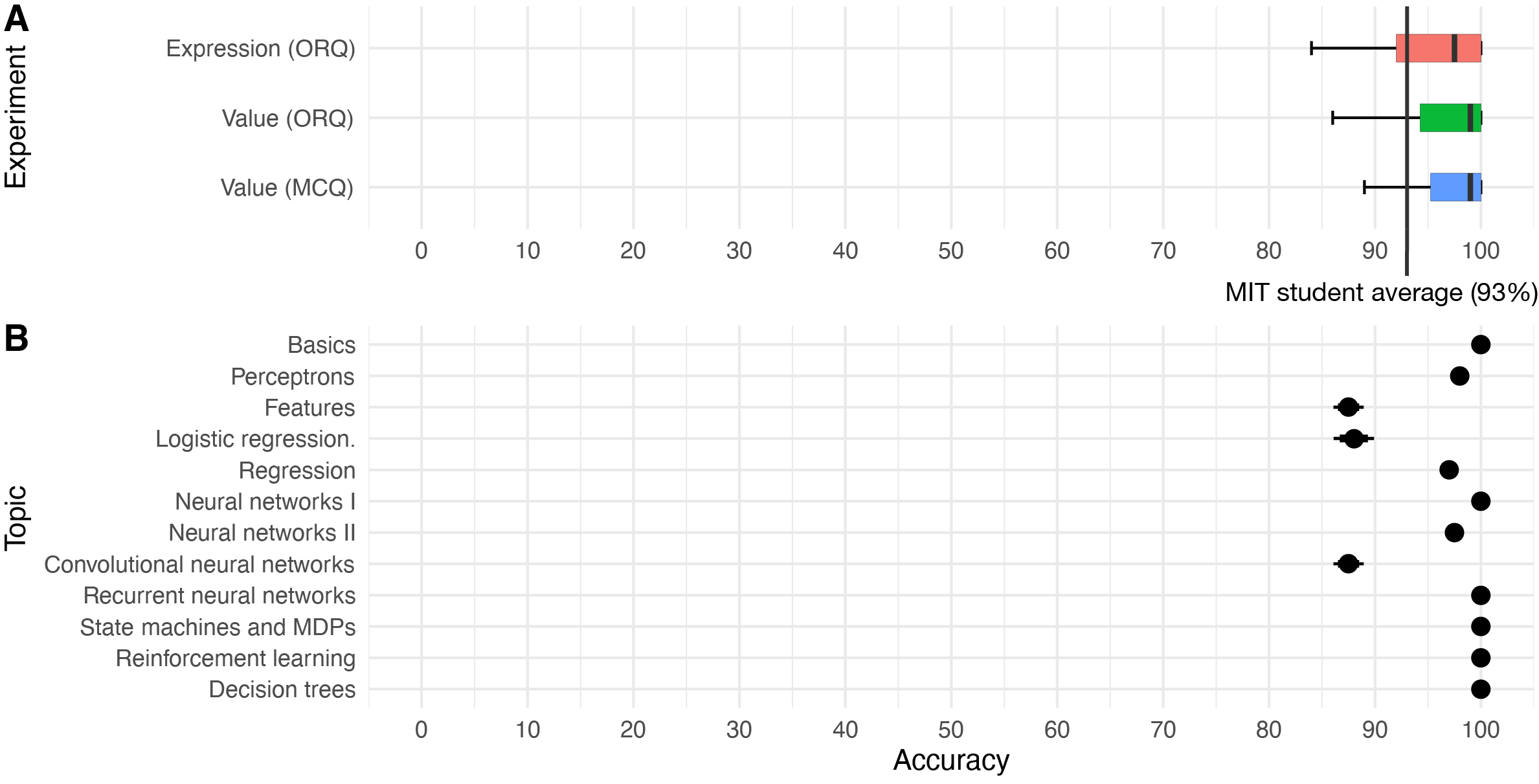}
    \caption{Accuracy distributions across \textbf{(A)} evaluation methods and \textbf{(B)} topics. The line overlay indicates the average student score.}
    \label{fig:results}
\end{figure}

We find that our model achieves grade A performance on its evaluation, when evaluated on both open-response questions and multiple-choice questions. A summary of the results are shown in Table \ref{tab:results} and Figure \ref{fig:results}.

\begin{figure}
    \centering
    \includegraphics[width=0.7\columnwidth]{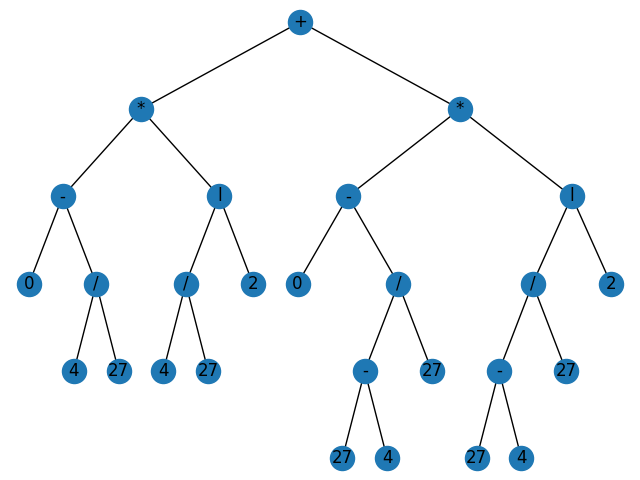}
    \caption{An automatically generated expression tree for the question ``What is the entropy of the left side of a region containing 27 points where the plane has 45 points in total and 4 points on the left are positive?" Note ``l" denotes the log operator.}
    \label{results:automatictrees}
\end{figure}

This is performed at super-human speed, answering each question in milliseconds, whereas humans typically take minutes to solve questions. Table \ref{results:examples} shows example questions and answers generated by our model for each topic. In the supplementary material, we provide 60 additional example questions and answers generated by our model, five for each of the course topics. Figure \ref{results:automatictrees} shows an example expression tree automatically generated by our model given a question. The supplementary material consists of 25 expression trees automatically generated by our model from questions in our dataset. 

\begin{table}
    \small
    \centering
    \begin{tabular}{|l|l|l|}
        \hline
        \textbf{Topic} & \textbf{Question} & \textbf{Answer} \\
        \hline \hline
        Basics & Compute the magnitude of $[10, 10, 1]$. & $14.18$ \\
        \hline
        Perceptrons & If the decision boundary of a classifier is $\theta$, where $\theta =$ & $4$ \\
        & $(4, 1)$, how does it classify point $p$, where $p = (2, -4)$? & \\
        \hline
        Features & A point $1$ has label $1$. Compute the margin of a classifier & $2$ \\
        & on this point. Let the $\theta$ of the classifier be $-1$ and the $\theta_0$ & \\
        & of the classifier be $-1$. & \\
        \hline
        Logistic & If we have $x = (0, -1)$, $\theta = (1, -2)$, and $\theta_0 = -3$, then & $-1$ \\
        Regression & what is the result of $\theta x + \theta_0$? & \\
        \hline
        Regression & If we let $\theta = 1$ and $\lambda = 0.5$, what is the mean squared error & $2.25$ \\
        & of the given points  $[(2, 0), (1, 1)]$? & \\
        \hline
        Neural & Neurons A and B take inputs $4$ and $2$ with weights $2$ and & $20.5$\\
        Networks I & $1$, respectively. Neuron A has offset $0.5$ and neuron B has & \\
        & offset $1$. Neuron C takes in the output of A and B with & \\
        & weights $1$ and $3$, respectively, and with offset $3$. What is & \\
        & the output? & \\
        \hline
        Neural & Compute the ReLU output of neuron C which takes the & $1.5$ \\
        Networks II & output of neuron A with weight $1$ and neuron B & \\
        & with weight $2$ and offset $1$. Neuron B has input $0$ and & \\
        & offset $1$. Neuron A has input $-1$ and offset $2$ with offset & \\
        & $0.5$. & \\
        \hline
        Convolutional & What is the minimum number of padding needed to & $8$ \\ 
        Neural & maintain the same output size if the input image is $50$ & \\
        Networks & by $50$ and the filter is $17$ by $17$? & \\
        \hline
        State Machines & Consider the input $x_t = [7, 14, 13, 10]$ to a state machine & $56$\\
        and Markov & with equations $s_t = f(s_{t-1}, x_t)$ and $y_t = g(s_t)$. Compute $y_4$ & \\
        Decision & if our initial conditions are $s_0 = 2$, $f(s_{t-1}, x_t) =$ & \\
        Processes & $\max(s_{t-1}, x_t)$, and $g(s_t) = 4s_t$. & \\
        \hline
        Reinforcement & Let $q = 2$. After Q learning, what is $q$ if $a = 0.3$ and $t = 8$? & $3.8$ \\
        Learning & & \\
        \hline
        Recurrent & Let $s_0 = 1.5$, $w = 1.5$, and $x = [1, 0, 2]$. Compute $s_3$ if $s_t =$ & $9.31$\\
        Neural & $w*s_{t-1} + x_t$. & \\
        Networks & & \\
        \hline
        Decision & Consider a 1D classification line on a 2D plane. There is a & $0.90$ \\
        Trees & total of $46$ points, $30$ of which are on the right and the & \\
        & rest on the left of the boundary. $5$ points on the left are & \\
        & classified positive. What is the entropy of the left region? & \\
        \hline
    \end{tabular}
    \caption{A sample question from each of the machine learning course topics and the answer generated by our model.}
    \label{results:examples}
\vspace{-24pt}
\end{table}

Table \ref{tab:input-output-examples_mc} shows a multiple-choice question, its associated answer choices, and the output produced by our model. For this example, the model is sampled 100 times and selects the highest probable answer from its 4 choices, where the correct answer is $1.5$. Our model finds the correct answer in the first attempt.

\begin{table}[h!]
    \small
    \centering
    \begin{tabular}{|l|l|l|l|}
        \hline
        \textbf{Topic} & \textbf{Question} & \textbf{Multiple Choice Options} & \textbf{Output} \\
        \hline
        \hline
        Regression & Let $1$ be the optimal $\theta$ by mean & $[-0.24, -2.24, -0.64, -53.9]$ & $[-0.24]$ \\
        & squared error. Given the data & & \\
        & points $[(0, 0), (1, -1), (2, y)]$ and & & \\ 
        & $\lambda = 1$, compute the value of $y$. & & \\
       
        \hline
    \end{tabular} 
    \caption{Multiple-choice question and answers. An example of the questions given to the model, the answers generated by the adversarial generator, and the output answers made by the model in order of the appearance. Note that the final value in output is the correct answer of the question, and the answering stops once the models achieves the correct answer. The correct answer is selected in the first attempt.}
    \label{tab:input-output-examples_mc}
\end{table}

Table \ref{tab:results_perceptrons} shows an example question and the resulting types of hints and three different types of hints generated by our model. This extension of the model to procedurally generate hints based on the solution equation allows it to provide insightful hints for learners struggling with the problem.

\begin{table}
\centering
\small
\begin{tabular}{|c|l|}
\hline
Question & A neural network has input $x_1$ with weight $w_1$ that goes into neuron $A$.\\
& Neuron $A$ also has input $\textit{OA}$ that has weight $w_{\textit{OA}}$. Neuron $C$ inputs\\ 
& the output of neuron $A$ with weight $w_{\textit{AC}}$. Neuron $C$ also has input $\textit{OC}$\\
& that has weight $w_{\textit{OC}}$. Neurons output the sum of  the products of each\\
& input with their respective weight. at has weight $w_{\textit{OA}}$. Neuron $C$ inputs\\ 
& the output of neuron $A$ with weight $w_{\textit{AC}}$. Neuron $C$ also has input $\textit{OC}$\\
& that has weigh $w_{\textit{OC}}$. Neurons output the sum of the products\\
& of each input with their respective weight. What is the output of\\
& neuron $C$ if $x_{1} = 2$, $w_{1} = 1$, $\textit{OA} = 0.5$, $w_{\textit{OA}} = 2$, $w_{\textit{AC}} = 1$, $\textit{OC} = 1$,\\
& and $w_{\textit{OC}} = 3$?\\
\hline
Answer & $(\textit{OA}*w_{\textit{OA}})*w_{\textit{AC}} + \textit{OC}*w_{\textit{OC}} = (0.5 * 2) * 1 + 1 * 3 = 4$\\
\hline
Expression & 1. $(\textit{OA}*w_{\textit{OA}})$\\
Hints & 2. $(\textit{OA}*w_{\textit{OA}})*w_{\textit{AC}}$\\
& 3. $\textit{OC}*w_{\textit{OC}}$\\
\hline
Value & 1. $0.5 * 2 = 1$\\
Hints & 2. $(0.5 * 2) * 1 = 1$\\
& 3. $1 * 3 = 3$\\
\hline
Example & For an example where $\textit{OA} = 8, w_{\textit{OA}} = 9, w_{\textit{AC}} = 3, \textit{OC} = 2, w_{\textit{OC}} = 4$:\\
Hints & 1. $8 * 9 = 72$\\
& 2. $(8 * 9) * 3 = 216$\\
& 3. $2 * 4 = 8$\\
\hline
\end{tabular}
\caption{Example questions and hints generated by our model.}
\label{tab:results_perceptrons}
\end{table}

\subsection{Baseline and Ablation Study}

To baseline the performance of our model, we use 120 open-response questions, with 10 questions per topic. We use a small number of questions since unlike our model which returns the numerical answer to the question, the responses of GPT-3 vary in length and content, thus requiring manual parsing of the responses given by GPT-3. We also perform an ablation study without the graphs and GNN such that the result of the Transformer feeds directly into the tree decoder.

We compare the results of these two models with that of GPT-3 tuned on machine learning in Table \ref{tab:baseline}. GPT-3 completely fails, achieving an overall value accuracy average of 7\%, while our original model achieves an overall value accuracy of 90\% and the reduced model achieves an overall value accuracy of 87\%. We present a detailed breakdown of the results comparing these three models by topic in Table \ref{tab:baseline}. To achieve these results, we train the models twice and take its averaged performance to mitigate randomness. These results highlight that although GPT-3 is able to generate text about concepts in Machine Learning, it completely fails in solving Machine Learning problems. This is expected, since GPT-3 has no understanding of simple topics in Linear Algebra and Calculus, such as the magnitude of a vector.

\begin{table}
    \small
    \centering
    \begin{tabular}{|c|l|c|c|c|}
    \hline
    \textbf{Week} & \textbf{Topic} & \textbf{Our Model} & \textbf{Our Model without GNN} & \textbf{GPT-3} \\
    \hline
    \hline
        1 & Basics & 1.00 & 0.50 & 0.00\\
        2 & Perceptrons & 1.00 & 0.95 & 0.20\\
        3 & Features & 0.65 & 0.50 & 0.00\\
        4 & Logistic regression & 0.70 & 0.50 & 0.20\\
        5 & Regression & 1.00 & 0.75 & 0.10\\
        6 & Neural networks I & 1.00 & 0.95 & 0.00\\
        7 & Neural networks II & 1.00 & 0.90 & 0.00\\
        8 & Convolutional neural networks & 0.90 & 0.95 & 0.00\\
        9 & Recurrent neural networks & 1.00 & 1.00 & 0.20\\
        10 & State machines and MDPs & 1.00 & 1.00 & 0.10\\
        11 & Reinforcement learning & 1.00 & 1.00 & 0.00\\
        12 & Decision trees & 1.00 & 0.90 & 0.00\\
        \hline
        & Overall average over topics & 0.94 & 0.83 & 0.07\\
        \hline
    \end{tabular}
    \caption{A comparison of the ORQ performance of our model vs. our model without GNN vs. GPT-3.}
    \label{tab:baseline}
\vspace{-20pt}
\end{table}

\subsection{Data Augmentation Performance}

We evaluate the performance of our data augmentation by varying the magnitude of the number of augmentations performed. For each topic, we downsample the number of produced augmentations, and we present the results in Table \ref{tab:data-augmentation-performance}. All topics generally decrease in performance as the number of augmentations decrease. However, we find topics which have many more questions to decay more gracefully than topics which contain fewer canonical questions. Certain topics may have spikes of higher performance at lower augmentations, which may occur due to the smaller solution space as the number of augmentations is decrease.d To account for randomness and variability within the model, the model is retrained four times, and we display the averaged results for each topic.

\begin{table}
    \small
    \centering
    \begin{tabular}{|c|l|c|c|c|c|c|c|}
    \hline
    & & \multicolumn{6}{c|}{\textbf{Augmentations per Question}} \\
    \cline{3-8}
    \textbf{Week} & \textbf{Topic} & \textbf{100} & \textbf{50} & \textbf{25} & \textbf{13} & \textbf{6} & \textbf{3} \\
    \hline
    \hline
        1 & Basics                        & 1.00 & 0.42 & 0.11 & 0.01 & 0.07 & 0.00\\
        2 & Perceptrons                   & 0.98 & 0.71 & 0.35 & 0.22 & 0.22 & 0.07\\
        3 & Features                      & 0.86 & 0.61 & 0.36 & 0.11 & 0.07 & 0.00\\
        4 & Logistic regression           & 0.86 & 0.76 & 0.12 & 0.02 & 0.00 & 0.02\\
        5 & Regression                    & 0.97 & 0.64 & 0.30 & 0.05 & 0.00 & 0.00\\
        6 & Neural networks I             & 1.00 & 0.86 & 0.34 & 0.01 & 0.05 & 0.06\\
        7 & Neural networks II            & 0.97 & 0.91 & 0.13 & 0.02 & 0.01 & 0.00\\
        8 & Convolutional neural networks & 0.86 & 0.91 & 0.54 & 0.12 & 0.06 & 0.04\\
        9 & Recurrent neural networks     & 1.00 & 0.61 & 0.23 & 0.00 & 0.05 & 0.00\\
        10 & State machines and MDPs     & 1.00 & 0.95 & 0.26 & 0.12 & 0.13 & 0.00\\
        11 & Reinforcement learning       & 1.00 & 1.00 & 0.64 & 0.00 & 0.02 & 0.00\\
        12 & Decision trees               & 1.00 & 0.80 & 0.60 & 0.00 & 0.00 & 0.00\\
        \hline
        & Overall average over topics & 0.96 & 0.77 & 0.33 & 0.06 & 0.06 & 0.02\\
        \hline
    \end{tabular}
    \caption{Performance of the model as the number of augmentations per question varies across all 12 topics. To remote randomness and variability, these values are averaged across the performance from retraining the model four times.}
    \label{tab:data-augmentation-performance}
\vspace{-20pt}
\end{table}

\subsection{Recursion}

We also demonstrate the capability of our system to solve questions which are recurrent in nature. We convert values to variables and allow for relationships across such variables. This is shown in the example in the Table, whose paraphrased template lays the foundation to allow for a variable depth recursive expression tree to be learned by the model. In the example, this is done by creating a relationship between the length of the input $x$ and the expected amount of recursive times needed to yield the solution, $s_{|\{x\}|}$. This process is done by our data augmentation on recurrent neural network questions, allowing for the model to learn the recursive nature of such problems, and as a result, data augmentation generates the data by which the model is trained to solve recursive problems. This is reflected in a score of 100\% on the recurrent neural networks topic in Table \ref{tab:results}.

\subsection{Limitations}
Despite demonstrating grade A performance on questions from MIT's Introduction to Machine Learning course, our model has several limitations. We provide examples of questions with incorrect answers in the supplementary material. A key component missing in our current model is the ability to comprehend questions involving pictorial components. In addition, our model is unable to write code for machine learning-related coding questions, which are present in MIT's Introduction to Machine Learning course. We are currently expanding upon the capabilities of our current system to address these limitations.

\section{Conclusion}
We present the first machine learning system capable of solving University undergraduate level Machine Learning questions with high accuracy, evaluated on both open-response questions and multiple-choice questions and demonstrating an overall grade A performance, and superseding MIT students. This work expands upon the current field of training machine learning models to learn courses, and we demonstrate the ability to create a model capable of solving machine learning problems by leveraging data augmentation techniques, such as templating and paraphrasing, a Transformer model, and a solution ranking procedure. In addition, we find our model capable of solving questions which are recurrent in nature. 

We also explore hint generation, opening the door to creating a teacher model which procedurally generates hints for learners. By leveraging the data augmentation component of our system with the hint generator, we are able to give students hints which support their understanding of the problem. This may automate labor intensive parts of running a course with minimal supervision. In addition, hint generation and providing methods of solving a problem demonstrates a much higher level of explainability compared to solely giving a numeric answer, which helps advance the area of explainable AI.

An extension of this work would be to expand upon the current size of training data, scaling up the data augmentation process to multiple courses. We are currently working on solving discrete math questions, specifically from MIT's Mathematics for Computer Science course, 6.042. In the future we would like to extend the set of questions handled to include coding questions and as well as question with visual components, using multi-skilled and multi-modal models.

\bibliography{bibliography}

\appendix

\section{Open Response Questions and Answers by our Model}

\subsection{Sample correct questions and model answers}

\begin{table}[!htb]
\centering
\small
\begin{tabular}{|l|c|}
\hline
\textbf{Basic Questions} & \textbf{Answer} \\ 
\hline \hline
1. Compute the magnitude of $[3, 12]$. & $12.37$ \\
\hline
2. If $x = [16, 4, 9]$, what is $||x||$? & $18.79$ \\
\hline
3. Find the Euclidean length of $[7, 0, 1]$ . & $7.07$ \\
\hline
4. Compute the magnitude of $[10, 10, 1]$. & $14.18$ \\
\hline
5. What is the magnitude of the vector $[0, 7]$? & $7$ \\
\hline
\end{tabular}
\caption{Example basic questions and answers generated by our model.}
\label{tab:results_basics1}
\end{table}

\begin{table}
\centering
\small
\begin{tabular}{|l|c|}
\hline
\textbf{Perceptrons Questions} & \textbf{Answer} \\ 
\hline \hline
1. If the decision boundary of a classifier is $\theta$, where $\theta = (4, 1)$, how & $4$ \\
does it classify point $p$, where $p = (2, -4)$?  & \\
\hline
2. How does a classifier with decision boundary $\theta$ classify a point $p$ if & $-16$ \\
$\theta = (2, 4)$ and $p = (0, -4)$? & \\
\hline 
3. What is the most number of mistakes made by the perceptron algorithm & $2704$ \\
if $13$ is the maximum magnitude of a point in the dataset and the dataset & \\
has a margin of $4$ to the separator. & \\
\hline 
4. Determine if the following two classifiers represent the same hyperplane, & $1$ \\
$[0, 1, 1]$ and $[0, 1, 1]$. If so, return $1$, and return anything else otherwise. & \\
\hline 
5. A classifier has a decision boundary where $\theta = (1, 0)$. What value does & $3$ \\
it classify $p$, where $p = (3, 0)$? & \\
\hline
\end{tabular}
\caption{Example perceptrons questions and answers generated by our model.}
\label{tab:results_perceptrons1}
\end{table}

\begin{table}
\centering
\small
\begin{tabular}{|l|c|}
\hline
\textbf{Features Questions} & \textbf{Answer} \\ 
\hline \hline
1. What is the margin of a classifier with $\theta = -1$ and $\theta_0 = -6$ on a & $-6$ \\
point $0$ with label $1$? & \\
\hline 
2. What is the loss for the data point $(0, -1)$ if we use NLL. Let $\theta = 2$ & $-0.69$ \\ and $\theta_0 = 0$. Also use natural log where the base is $2.71828$. & \\
\hline 
3. A point $1$ has label $1$. Compute the margin of a classifier on this & $2$ \\
point. Let the $\theta$ of the classifier be $-1$ and the $\theta_0$ of the classifier be $-1$. & \\
\hline
4. Given the values for $\theta = 2$ and $\theta_0 = 1$, compute the NLL loss on the & $-0.31$ \\
data point $(-1, 0)$. Use log base $e$ of $2.71828$ for the log. & \\
\hline
5. What does the sigmoid function return when you pass into it $1$? Hint: & $0.73$ \\
have $e = 2.71828$. & \\
\hline
\end{tabular}
\caption{Example feature questions and answers generated by our model.}
\label{tab:results_features}
\end{table}

\begin{table}
\centering
\small
\begin{tabular}{|l|c|}
\hline
\textbf{Logistic Regression Questions} & \textbf{Answer} \\ 
\hline \hline
1. What is the result of $\theta x + \theta_0$ if $x = (-1, 0)$, $\theta = (0, -1)$, and $\theta_0 = 3$? & $3$ \\
\hline
2. Let $\theta = (1, -1)$, $\theta_0 = 2$, and $x = (-1, 0)$. Compute $\theta x + \theta_0$. & $1$ \\
\hline
3. If you let $\theta = 1$ and $\eta = 0.05$, what is the updated $\theta$ value after one & $1$ \\
gradient descent step if the loss function is given by $( 0 \times \theta + 3)^2$ ? & \\
\hline
4. If we have $x = (0, -1)$, $\theta = (1, -2)$, and $\theta_0 = -3$, then what is the & $-1$ \\
result of $\theta x + \theta_0$? & \\
\hline
5. What is the value of $\theta x + \theta_0$ if $x = (-1, 0)$, $\theta = (-1, 0)$, and $\theta_0 = -3$? & $-2$ \\
\hline
\end{tabular}
\caption{Example logistic regression questions and answers generated by our model.}
\label{tab:results_logistic_regression}
\end{table}

\begin{table}
\centering
\small
\begin{tabular}{|l|c|}
\hline
\textbf{Regression Questions} & \textbf{Answer} \\ 
\hline \hline
1. If $f(\theta) = (3 \times \theta + 3)^{2}$ and $\theta = 4$ what is $f(\theta)$? & $225$\\
\hline
2. Given $\theta = 3$ and $\lambda = 1$, compute the mean squared error with the data & $19.5$ \\
points $[(2, 0), (2, 5)]$. & \\
\hline
3. With $\lambda = 1$, the optimal $\theta = 1$. If the data points are $[(0, 1), (1, 2),$ & $0.27$ \\
$(2, y)]$, what is the value of $y$? The optimal $\theta$ is computed by mean & \\
squared error. & \\
\hline
4. If we let $\theta = 1$ and $\lambda = 0.5$, what is the mean squared error of the given & $2.25$ \\
points  $[(2, 0), (1, 1)]$? & \\
\hline
5. If $f(\theta)$ is $(7 \times \theta + 8)^2$, what is $f(\theta)$ when $\theta =  15.4$? & $13409.64$ \\
\hline
\end{tabular}
\caption{Example regression questions and answers generated by our model.}
\label{tab:results_regression}
\end{table}

\begin{table}
\centering
\small
\begin{tabular}{|l|c|}
\hline
\textbf{Neural Networks I Questions} & \textbf{Answer} \\ 
\hline
\hline
1. Neurons A and B take inputs $4$ and $2$ with weights $2$ and $1$, respectively. & $20.5$\\
Neuron A has offset $0.5$ and neuron B has offset $1$. Neuron C takes in the & \\
output of A and B with weights $1$ and $3$, respectively, and with offset $3$. & \\
What is the output? & \\
\hline
2. Neuron C is the output neuron and neuron A takes the input. Compute & $11$\\
the output with the given architecture and inputs. Neuron C takes in the & \\
offset value $oC = 4$ with weight $wOC = 2$. Neuron C also takes in the & \\
neuron A with weight $wAC = 1$. Neuron A takes in the input value $x1 = 3$ & \\
output of with weight $w1 = 1$ and offset value $oA = 0.5$ and offset weight & \\
$wOC = 2$. & \\
\hline
3. In a fully-connected feed-forward network, how many weights (including & $3200$ \\
biases) are there for one layer with $40$ inputs and $40$ outputs? & \\
\hline
4. A fully-connected neural network has $90$ outputs and $40$ inputs. How & $7200$ \\
many total weights are there including the biases? & \\
\hline
5. Neuron A and Neuron C are the input and output neurons of a neural & $8$ \\
network. Neuron A takes in value $x1 = 2$ with weight $w1 = 1$ and offset & \\
value $oA = 0.5$ with weight $wOA = 4$. Neuron C takes in the output of & \\
neuron A with weight $wAC = 1$ and offset value $oC = 2$ with weight & \\
$wOC = 2$. What is the output of neuron C? & \\
\hline
\end{tabular}
\caption{Example neural network questions and answers generated by our model.}
\label{tab:results_neural_networks_1}
\end{table}

\begin{table}
\centering
\small
\begin{tabular}{|l|c|}
\hline
\textbf{Neural Networks II Questions} & \textbf{Answer} \\ 
\hline
\hline
1. A ReLU is applied to the output of neuron C, which takes in outputs & $15.5$ \\
from neurons A with weight $wAC = 1$ and B with weight $wBC = 2$ and & \\
offset $oC = 1$. Neuron A takes in value $x1 = 4$ with weight $w1 = 1$ and & \\
offset value $oA = 0.5$. Neuron B takes in input $x2 = 4$ with an offset of $1$. & \\
\hline
2. A neural network has inputs $x1 = 4$ with weight $2$ and $x2 = 2$ with & $16.5$ \\
weight $1$ and offset value $oA = 0.5$. Neuron B inputs $x2$ with offset $1$. & \\
Neuron C takes in the output of neurons A and B with offsets $wAC = 1$ & \\
and $wBC = 2$, respectively. Neuron C has offset value $oC = 2$ and & \\
applies a ReLU on its output. Compute the output. & \\
\hline
3. Neuron C is the output neuron which applies a ReLU on its output and & $6$ \\
neuron A is the input neuron to a neural network. Compute the output of a & \\
neural network with the given architecture and inputs. Neuron C takes in & \\
the offset value $oC = 2$ with weight $wOC = 3$. Neuron C takes in the output& \\
of neuron A with weight $wAC = 1$. Neuron A takes in the input value & \\
$x1 = 0$with weight $w1 = 2$ and offset value $oA = 0.5$ and offset weight & \\
$wOC = 3$. & \\
\hline
4. Neurons A and B take inputs $2$ and $1$ with weights $2$ and $1$, respectively. & $10.5$ \\
Neuron A has offset $0.5$ and neuron B has offset $1$ . Neuron C takes in the & \\
output of A and B with weights $1$ and $2$, respectively, and with offset 2. & \\
Neuron C also applies a ReLU on its output. What is the output? & \\ 
\hline
5. Compute the ReLU output of neuron C which takes the output of neuron & $1.5$ \\
A with weight $1$ and neuron B with weight $2$ and offset $1$. Neuron B has & \\
input $0$ and offset $1$. Neuron A has input $-1$ and offset $2$ with offset $0.5$. & \\
\hline
\end{tabular}
\caption{Example neural network questions and answers generated by our model.}
\label{tab:results_neural_networks_2}
\end{table}

\begin{table}
\centering
\small
\begin{tabular}{|l|c|}
\hline
\textbf{Convolutional Neural Networks Questions} & \textbf{Answer} \\ 
\hline
\hline
1. An image $I$ has length $5$ and filter $F$ has length $2$, what is the length & $4$\\
of the result of applying $F$ to $I$? & \\
\hline
2. The row of an image $[1, 1, 1]$ has a filter $[0, 3, 0]$ applied to it. What is & $3$ \\
the resulting value if they both align? & \\
\hline
3. What is the minimum number of padding needed to maintain the same & $8$ \\ 
output size if the input image is $50$ by $50$ and the filter is $17$ by $17$? & \\
\hline
4. Given that there are $22$ inputs to a zero-padded max pooling layer and & $11$ \\
a stride length of $2$, compute the number of output units if we also know the & \\
pooling filter size of $3$? & \\
\hline
5. What is the length of the output when we use an image of length $52$ & $24$ \\
and a filter of length $5$ if we use a stride length of $2$? & \\
\hline
\end{tabular}
\caption{Example convolutional neural network questions and answers generated by our model.}
\label{tab:results_convolutional_neural_networks}
\end{table}

\begin{table}
\centering
\small
\begin{tabular}{|l|c|}
\hline
\textbf{Recurrent Neural Networks Questions} & \textbf{Answer} \\ 
\hline
\hline
1. Consider a very simple RNN, defined by the following equation: & $0.525$\\
$s_t = w*s_{t-1} + x_t$. Given $s_0 = 0$, $w = 0.1$, and $x = [0.25, 0.5]$, & \\
what is $s_2$? & \\
\hline
2. An RNN is defined as $s_t = w*s_{t-1} + x_t$. If $s_0 = 1$, $w = 1$, and & $5$\\
$x = [2, 2, 0]$, what is $s_3$? & \\
\hline
3. What is the RNN result $s_2$ if $s_0 = 2$, $w = 0$, and $x = [9, 5]$ if & $5$\\
we let $s_t = w*s_{t-1} + x_t$?  & \\
\hline
4. We define an RNN as $s_t = w*s_{t-1} + x_t$. What is $s_2$ if $s_0 = 3$, & $3.75$\\
$w = 0.5$, and $x = [2, 2]$?  & \\
\hline
5. Let $s_0 = 1.5$, $w = 1.5$, and $x = [1, 0, 2]$. Compute $s_3$ if & $9.31$\\
$s_t = w*s_{t-1} + x_t$. & \\
\hline
\end{tabular}
\caption{Example recurrent neural network questions and answers generated by our model.}
\label{tab:results_recurrent_neural_networks}
\end{table}

\begin{table}
\centering
\small
\begin{tabular}{|l|c|}
\hline
\textbf{State Machines and MDP Questions} & \textbf{Answer} \\ 
\hline
\hline
1. Let a state machine be described with the equations $s_t = $ & $18$\\
$f(s_{t-1}, x_t)$ and $y_t = g(s_t)$, where $x_t$ is the input. If $s_0 = 6$, &\\
$f(s_{t-1}, x_t) = \max(s_{t-1}, x_t)$, and $g(s_t) = s_t$, what is the output &\\
$y_3$ after the inputs $[16, 9, 18]$? & \\
\hline
2. If we have a state machine, defined as $s_t = f(s_{t-1}, x_t)$ and & $0$\\
$y_t = g(s_t)$, where $x_t$ is the input, what is the output $y_3$ &\\
if we have $s_0 = 14$, $f(s_{t-1}, x_t) = \max(s_{t-1}, x_t)$, $g(s_t) = 0s_t$, & \\
and we input $[14, 15, 9]$? & \\
\hline
3. Consider the input $x_t = [7, 14, 13, 10]$ to a state machine with & $56$\\
equations $s_t = f(s_{t-1}, x_t)$ and $y_t = g(s_t)$. Compute $y_4$ &\\
if our initial conditions are $s_0 = 2$, $f(s_{t-1}, x_t) = \max(s_{t-1}, x_t)$, & \\
and $g(s_t) = 4s_t$. & \\
\hline
4. A state machine is defined by the equations $s_t = f(s_{t-1}, x_t)$ & $51$\\
and $y_t = g(s_t)$. Given the conditions $s_0 = 7$, $f(s_{t-1}, x_t) = $ & \\
$\max(s_{t-1}, x_t)$, and $g(s_t) = 3s_t$, compute $y_5$ if the input is & \\
$x_t = [17, 4, 14, 2, 16]$. & \\
\hline
5. What is the output $y_5$ of a state machine with equations & $76$ \\
$s_t = f(s_{t-1}, x_t)$ and $y_t = g(s_t)$, conditions $s_0 = 10$, $f(s_{t-1}, x_t) = $ &\\
$\max(s_{t-1}, x_t)$, and $g(s_t) = 4s_t$, and input $x_t = [19, 17, 5, 9, 18]$? &\\
\hline
\end{tabular}
\caption{Example state machines and MDP questions and answers generated by our model.}
\label{tab:results_state_machines_mdps}
\end{table}

\begin{table}
\centering
\small
\begin{tabular}{|l|c|}
\hline
\textbf{Reinforcement Learning Questions} & \textbf{Answer} \\
\hline
\hline
1. What is the updated Q value of a tuple (s, a) if $q = 7$, the $a = 0.1$, & $6.9$\\
and $t = 6$? & \\
\hline
2. If $q = 8$, what is its updated value after applying Q learning if & $7.6$\\
$a = 0.1$ and $t = 4$? & \\
\hline
3. Let $q = 2$. After Q learning, what is $q$ if $a = 0.3$ and $t = 8$? & $3.8$\\
\hline
4. If $a = 0.4$ and $t = 10$, what is the Q learning value after applying & $7.6$\\
one tuple (s, a) if $q = 6$?  & \\
\hline 
5. After applying Q learning to $q = 4$, what is its value? Let the & $8.8$\\
$t = 10$ and $a = 0.8$. & \\
\hline
\end{tabular}
\caption{Example reinforcement learning questions and answers generated by our model.}
\label{tab:results_reinforcement_learning}
\end{table}

\begin{table}[!htb]
\centering
\small
\begin{tabular}{|l|c|}
\hline
\textbf{Decision Trees and Nearest Neighbors Questions} & \textbf{Answer} \\ 
\hline
\hline
1. A left region has $6$ points classified as positive. There are $44$ points in & $0.71$ \\
the plane, and $31$ points on the left. Compute the entropy. & \\
\hline 
2. Consider a 1D classification line on a 2D plane. There is a total of $46$ & $0.90$ \\
points, $30$ of which are on the right and the rest on the left of the & \\
$5$  boundary. points on the left are classified positive. What is the entropy & \\
of the left region? & \\
\hline 
3. If a region has $24$ points on the left and 46 points total. $8$ points that & $0.92$ \\
are on the left are positive. Compute the entropy. & \\
\hline 
4. The left side of a region has $26$ points. Of the $26$ points, $4$ are & $0.62$ \\
classified as positive. What is the entropy of the left region if there are $45$ & \\
points in total? & \\
\hline 
5. What is the entropy of the left side of a region containing $30$ points & $0.83$ \\
where the plane has $47$ points in total and $8$ points on the left are & \\
positive? & \\
\hline
\end{tabular}
\caption{Example decision trees and nearest neighbor questions and answers generated by our model.}
\label{tab:results_decision_trees_nn}
\end{table}

\newpage

\subsection{Expression Trees generated by our model for open response questions.}

\begin{figure}[!htb]
    \centering
    \includegraphics[width=0.5\columnwidth]{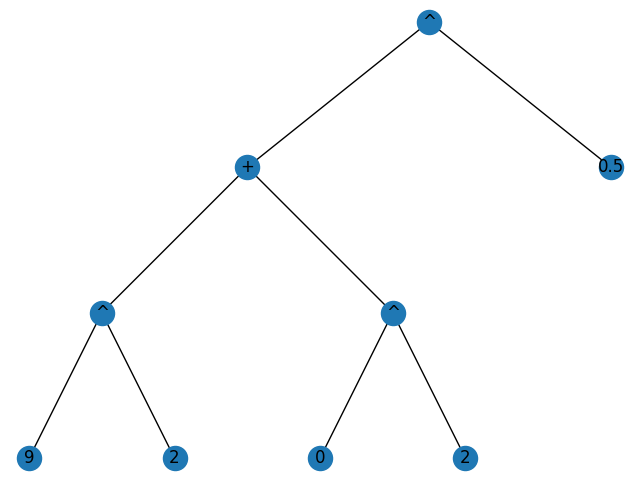}
    \caption{Basics: $(x^2 + y^2)^{0.5} = (9^2 + 0^2)^{0.5}$}
    \label{fig:t1}
\end{figure}

\begin{figure}
    \centering
    \includegraphics[width=0.5\columnwidth]{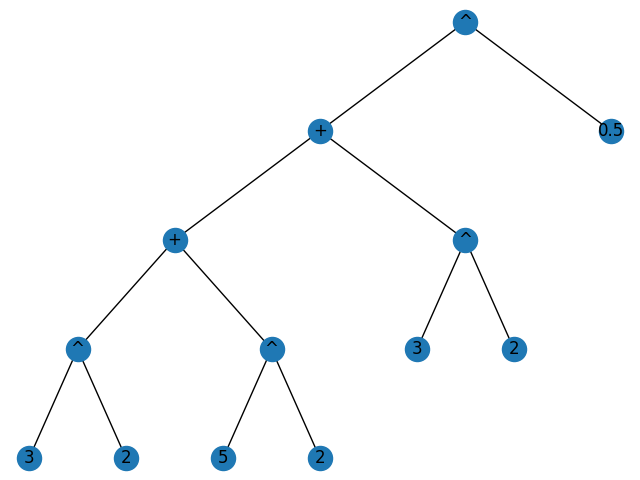}
    \caption{Basics: $(x^2 + y^2 + z^2)^{0.5} = (3^2 + 5^2 + 3^2)^{0.5}$}
    \label{fig:t2}
\end{figure}

\begin{figure}
    \centering
    \includegraphics[width=0.5\columnwidth]{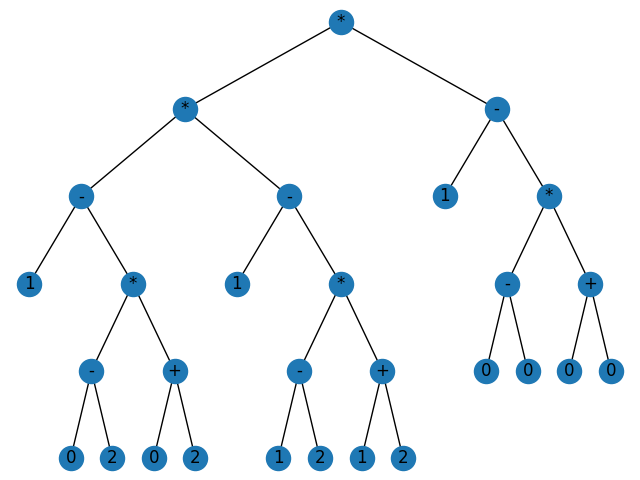}
    \caption{Perceptrons: $(a -(b-c)*(d+p))*(f-(g-h)*(i+j))*(k-(l-m)*(n+o)) = ((1-(0-2)*(0+2))*(1-(1-2)*(1+2)))*(1-(0-0)*(0+0))$}
    \label{fig:t3}
\end{figure}

\begin{figure}
\centering
\includegraphics[width=0.5\columnwidth]{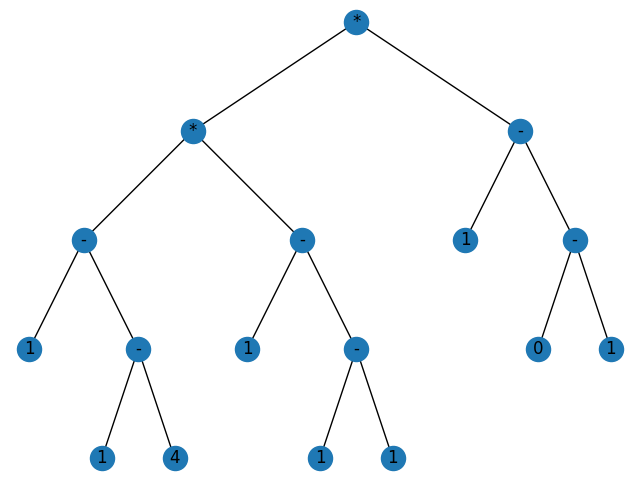}
\caption{Perceptrons: $((a-(b-c))*(d-(e-f)))*(g-(h-i)) = ((1-(1-4))*(1-(1-1)))*(1-(0-1))$}
\label{fig:t4}
\end{figure}

\begin{figure}
    \centering
    \includegraphics[width=0.5\columnwidth]{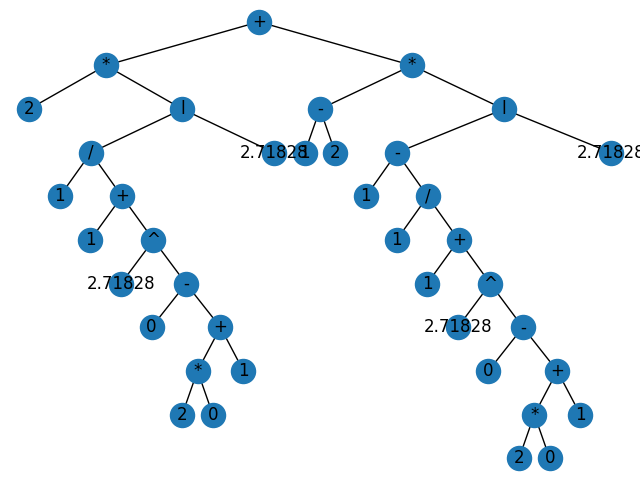}
    \caption{Features: $s*\ln \frac{1}{1+e^{(m-(n*o+p)}} + (q-r)*(1- \ln \frac{1}{1+e^{(m-(n*o+p)}}) = 2*\ln \frac{1}{1+e^{(0-(2*0+1)}} + (1-2)*(1- \ln \frac{1}{1+e^{(0-(2*0+1)}})$}
    \label{fig:t5}
\end{figure}

\begin{figure}
\centering
\includegraphics[width=0.5\columnwidth]{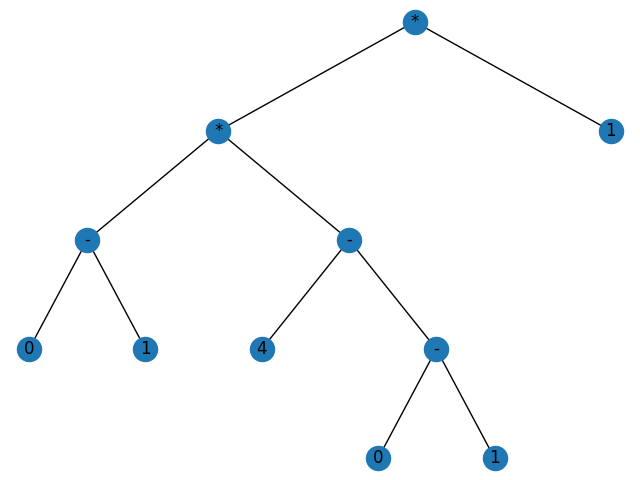}
\caption{Features: $(u-v)*(v-(w-x))*y = ((0-1)*(4-(0-1))*1$ }
\label{fig:t6}
\end{figure}

\begin{figure}
    \centering
    \includegraphics[width=0.5\columnwidth]{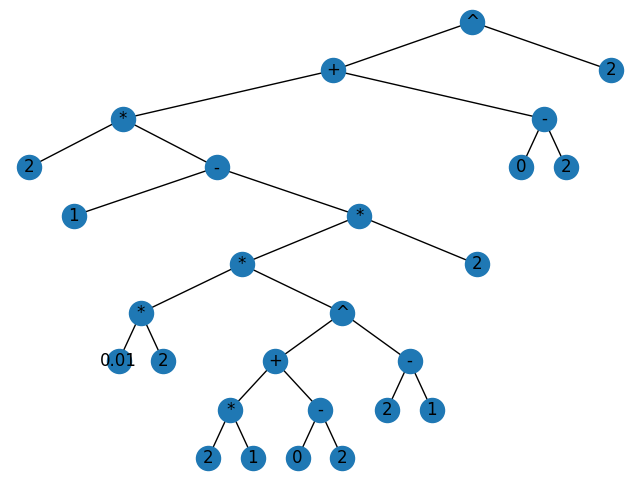}
    \caption{Logistic regression: $(f*(g-((h*i)*(j*k)+(l-m)^{n-o})*p))+(q-r)))^2
    = (2*(1-((0.01*2)*(2*1)+(0-2)^{2-1})*2))+(0-2)))^2$}
    \label{fig:t7}
\end{figure}

\begin{figure}
    \centering
    \includegraphics[width=0.5\columnwidth]{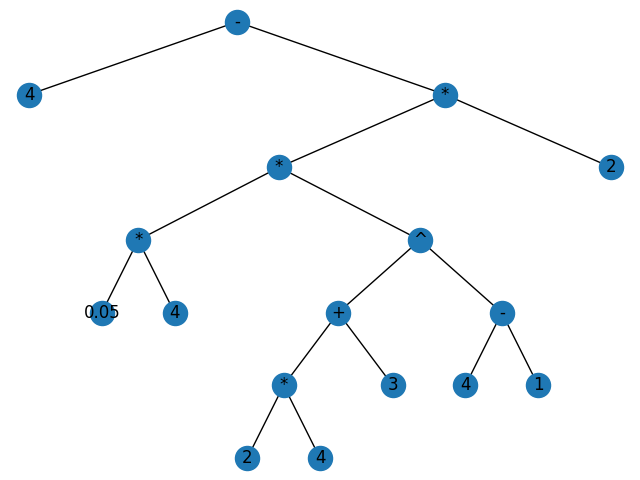}
    \caption{Logistic regression: $ g - (h*i) *((k*l)+m)^{n-o})*p 
    = 4 - (0.05*4) *((2*4)+3)^{4-1})*2$}
    \label{fig:t8}
\end{figure}

\begin{figure}
    \centering
    \includegraphics[width=0.5\columnwidth]{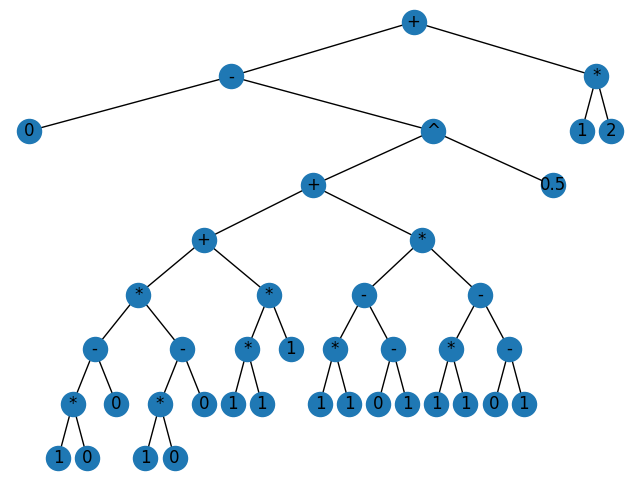}
    \caption{Regression: $(0-((a*b)-c)*((d*e)-f)+((g*h)*i)+(((j*k)-(l-m))*((n*o)-(p-q)))^{0.5})+(r*s) = (0-((1*0)-0)*((1*0)-0)+((1*1)*1)+(((1*1)-(0-1))*((1*1)-(0-1)))^{0.5})+(1*2)$}
    \label{fig:t9}
\end{figure}

\begin{figure}
    \centering
     \includegraphics[width=0.5\columnwidth]{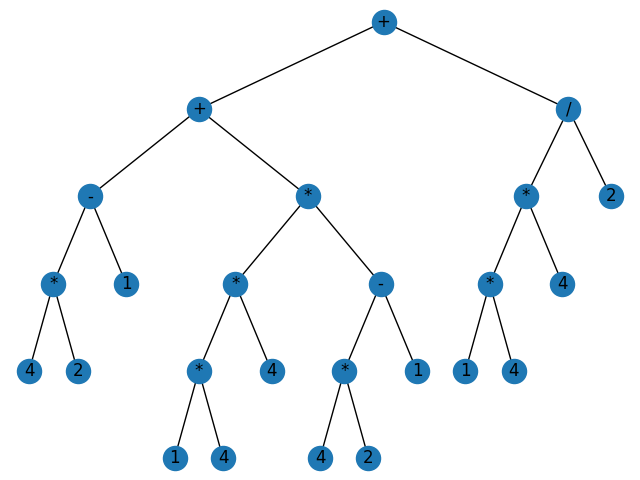}
    \caption{Regression: $((a*b)-c)+(((d*e)*f)*((g*h)-i))+(((j*k)*l)/m) = ((4*2)-1)+(((1*4)*4)*((4*2)-1))+(((1*4)*4)/2)$}
    \label{fig:t10}
\end{figure}

\begin{figure}
    \centering
     \includegraphics[width=0.5\columnwidth]{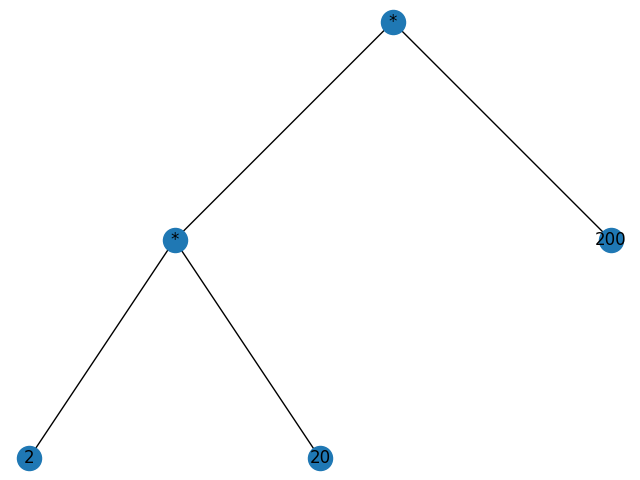}
    \caption{$(x*y)*z = (2*20)*200$}
    \label{fig:t11}
\end{figure}

\begin{figure}
    \centering
     \includegraphics[width=0.5\columnwidth]{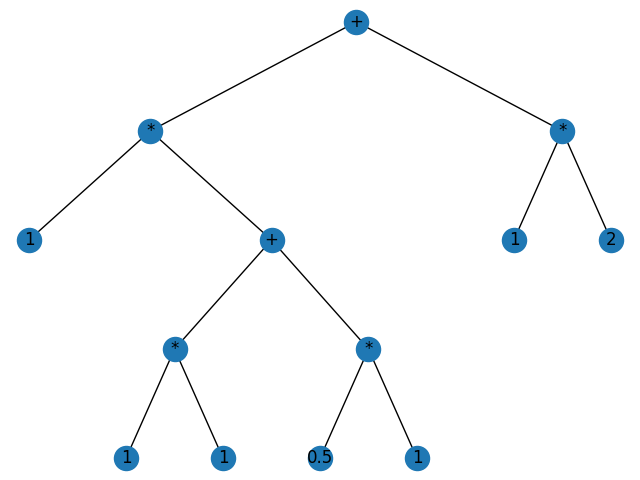}
    \caption{Neural networks I: $(u*((v*w)+(w*x))) + (y*z) = (1*((1*1)+(0.5*1)))+1*2$}
    \label{fig:t12}
\end{figure}

\begin{figure}
    \centering
     \includegraphics[width=0.5\columnwidth]{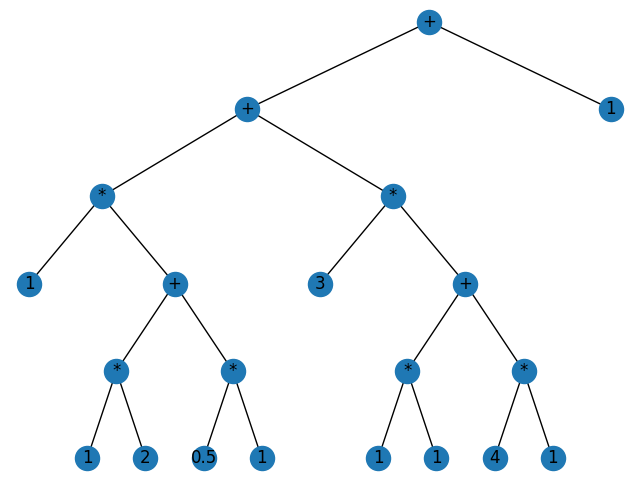}
    \caption{Neural networks I: $p*(q*r+ s*t) + u * (v*w + x*y) + z  = (1*(1*2+0.5*1))+3*(1*1+4*1)+ 1$}
    \label{fig:t13}
\end{figure}

\begin{figure}
    \centering
     \includegraphics[width=0.5\columnwidth]{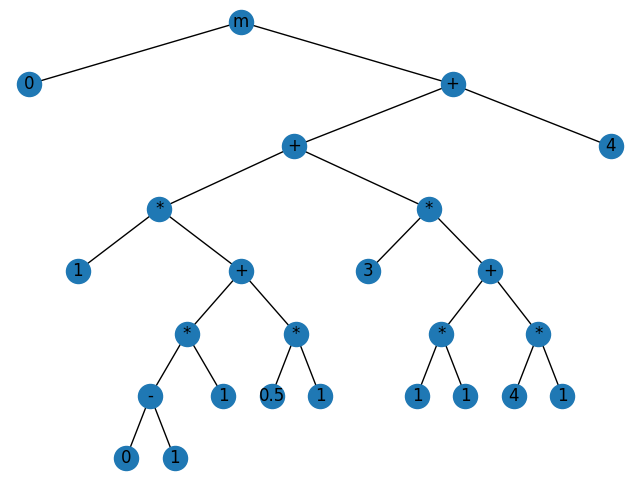}
    \caption{Neural networks II: $\max(0, (g*(h-i)*j+k*l)+m*(n*o+p*q)+r)  = \max(0, (1*(0-1)*1+0.5*1)+3*(1*1+4*1)+4)$}
    \label{fig:t14}
\end{figure}

\begin{figure}
    \centering
     \includegraphics[width=0.5\columnwidth]{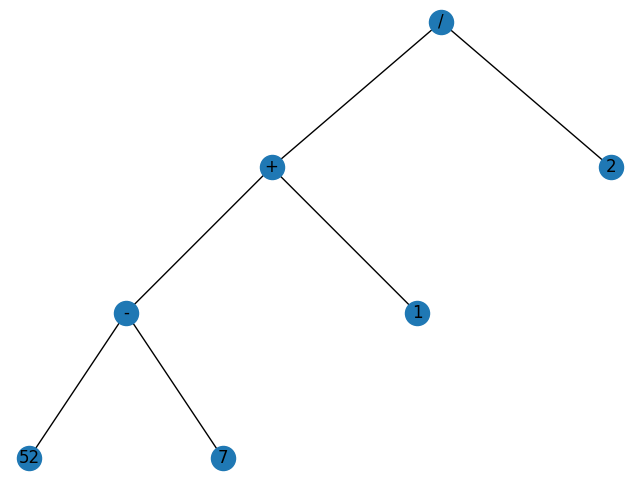}
    \caption{Convolutional neural networks: $((a-b)+c)/2  = ((52-7)+1)/2$}
    \label{fig:t15}
\end{figure}

\begin{figure}
    \centering
     \includegraphics[width=0.5\columnwidth]{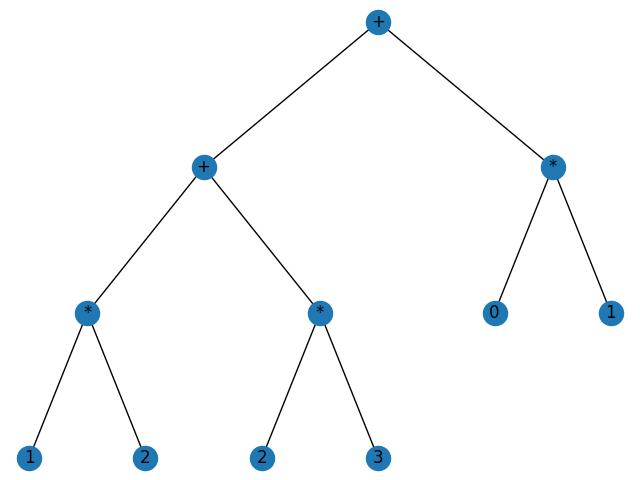}
    \caption{Convolutional neural networks: $(a*b+c*d) +(g*h) = (1*2 + 2*3) + (0*1)$}
    \label{fig:t16}
\end{figure}

\begin{figure}
    \centering
     \includegraphics[width=0.5\columnwidth]{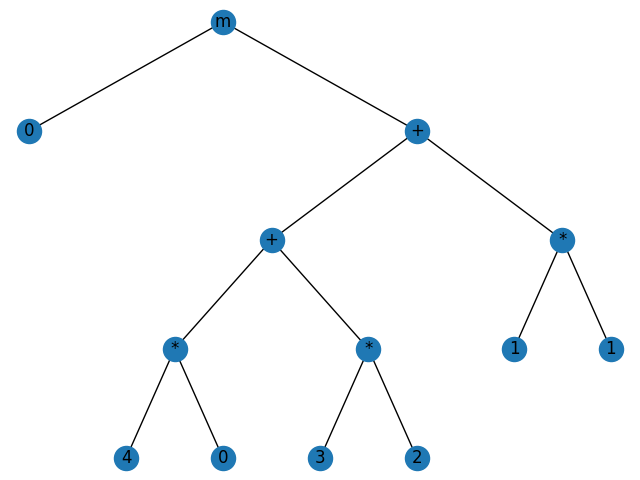}
    \caption{Convolutional neural networks: $\max(0, (p*q + r*s)+(t*u)) = \max(0, (4*0+3*2) + (1*2))$}
    \label{fig:t17}
\end{figure}

\begin{figure}
    \centering
    \includegraphics[width=0.5\columnwidth]{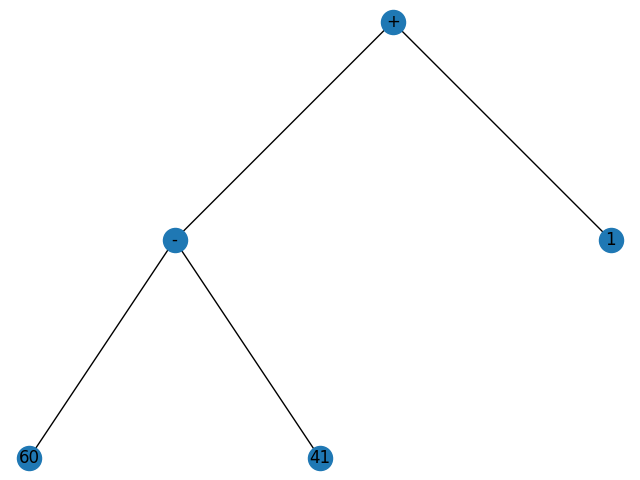}
    \caption{Convolutional neural networks: $(a-b) + c = (60-41) + 1$}
    \label{fig:t18}
\end{figure}

\begin{figure}
    \centering
    \includegraphics[width=0.5\columnwidth]{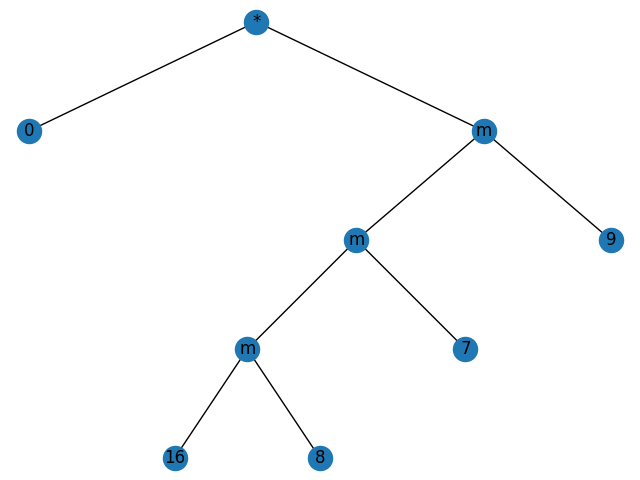}
    \caption{State machines and MDPs: $c*\max(\max(\max(m, n), o), p) = 0*\max(\max(\max(16, 8), 7), 9) $}
    \label{fig:t19}
\end{figure}

\begin{figure}
    \centering
    \includegraphics[width=0.5\columnwidth]{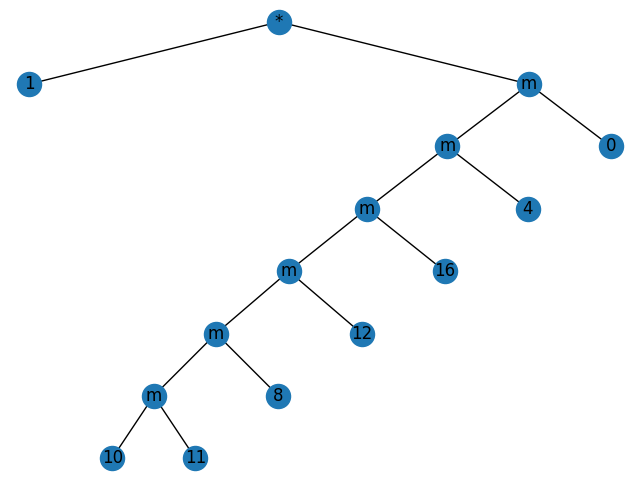}
    \caption{State machines and MDPs: $f \max(\max(\max(\max(\max(g, h), i), j), k), l), m)=1*\max(\max(\max(\max(\max(10, 11), 8), 12), 16), 4), 0)$}
    \label{fig:t20}
\end{figure}

\begin{figure}
    \centering
    \includegraphics[width=0.5\columnwidth]{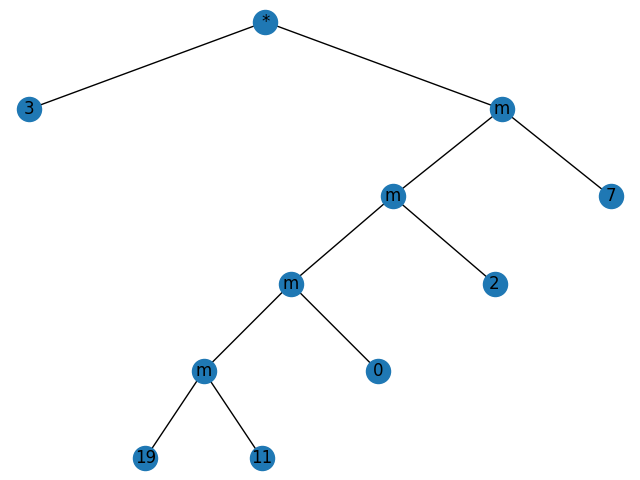}
    \caption{State machines and MDPs: $a\max(\max(\max(\max(p, q), r), s), t) = 3*\max(\max(\max(\max(19, 11), 0), 2), 7)$}
    \label{fig:t21}
\end{figure}

\begin{figure}
    \centering
    \includegraphics[width=0.5\columnwidth]{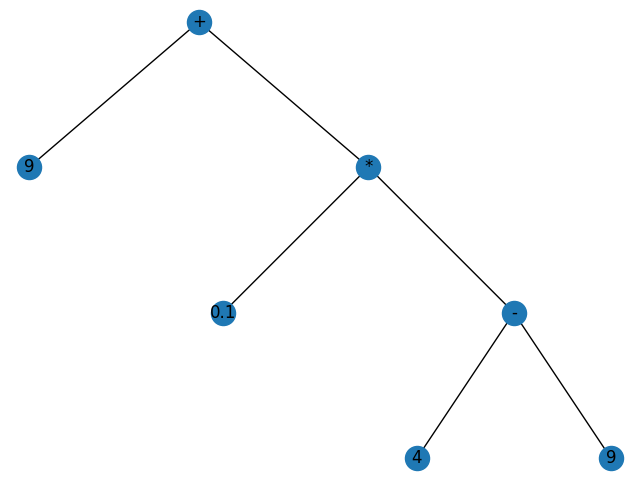}
    \caption{Reinforcement learning: $a + b * (c -d) = 9 + 0.1 * (4-5)$}
    \label{fig:t22}
\end{figure}

\begin{figure}
    \centering
    \includegraphics[width=0.5\columnwidth]{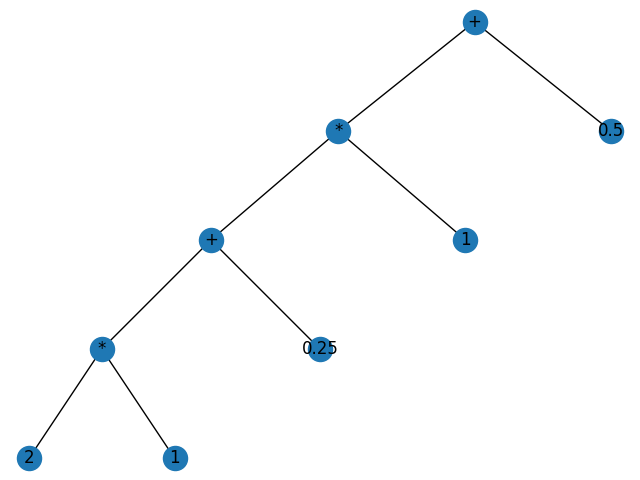}
    \caption{Recurrent neural networks: ($(a*b + c)*d + e)*f = ((2*1 + 0.25)*1 + 0.5) $ }
    \label{fig:t23}
\end{figure}

\begin{figure}
    \centering
    \includegraphics[width=0.5\columnwidth]{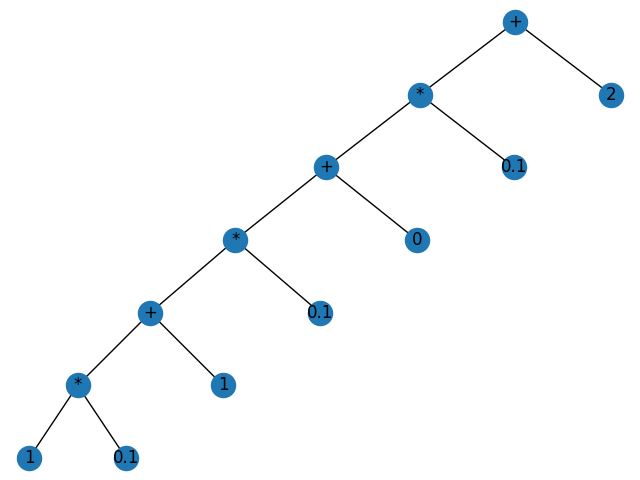}
    \caption{Recurrent neural networks: $(a*b + c)*d + e)*f+g = ((1*0.1 + 1)*0.1 + 0)*0.1+2 $ }
    \label{fig:t24}
\end{figure}

\begin{figure}
    \centering
    \includegraphics[width=0.5\columnwidth]{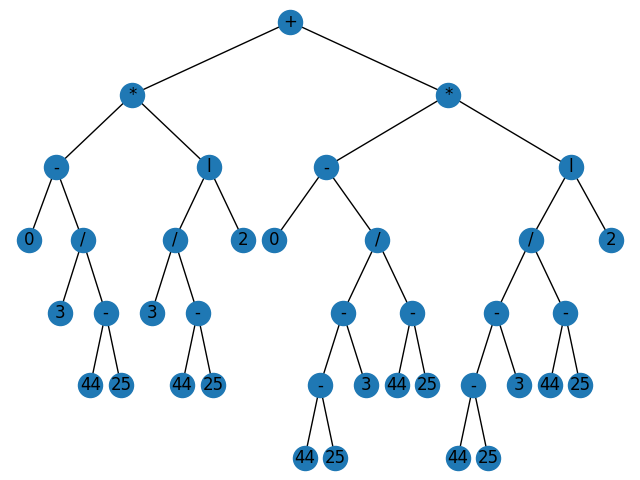}
    \caption{Decision trees: $(0 - a / (b-c))*\log_2(a/(b-c)) + (0 - (b-c-a)/(b-c))*\log_2((b-c-a)/(b-c)) = (0 - 3 / (44-25) * \log_2(3/(44-25)) + (0 - (44 - 25 -3)/(44-25)*\log_2((44 - 25 -3) / (44-25))$}
    \label{fig:t25}
\end{figure}

\clearpage
\section{Multiple Choice Questions and Answers by our Model}

\begin{table}[!htb]
    \centering
    \begin{tabular}{|p{3cm}|p{7cm}|l|}
        \hline
        \multicolumn{1}{|c}{Topic} & \multicolumn{1}{|c|}{Question/Multiple Choice Options} & \multicolumn{1}{c|}{Our Model's Answers} \\
        \hline
        \hline
        Basics & \specialcell{\textbf{Question:} What is the magnitude of the \\ vector $[1, 6, 4]$? \\ \textbf{Answer Options:} $[8, 4.12, 2.38, 1.41]$} & $[1.41]$ \\
        \hline
        Perceptrons & \specialcell{\textbf{Question:} Do the two classifiers $[0, 1, 0]$ \\ and $[2, 2, 0]$ represent the same \\ hyperplane? Return $1$ if true and \\ anything else otherwise. \\ \textbf{Answer Options:} $[230.0, -40.0, 23.2, $ \\ $20.0]$} & $[-40, 23.2, 20]$ \\
        \hline
        Features & \specialcell{\textbf{Question:} Consider the point $(0, 2)$, the \\ $\theta = 2$ and the $\theta_0 = 1$. What is the NLL \\ loss? Use natural log, where the base is \\ $2.71828$. \\ \textbf{Answer Options:} $[-0.02, 0.69, 2.06,$ \\ $1.06]$} & $[0.69]$ \\
        \hline
        Logistic regression & \specialcell{\textbf{Question:} Given a function $(2 \theta - 2)^2$, \\ calculate the value of the function after \\ one gradient descent update if $\theta = 1$ and \\ $\eta = 0.01$. \\ \textbf{Answer Options:} $[\infty, 0.1, 0.41, 0]$} & $[0.1, 0.41, 0]$ \\
        \hline
        Regression & \specialcell{\textbf{Question:} Let $1$ be the optimal $\theta$ by \\ mean squared error. Given the \\ datapoints $[(0, 0), (1, -1), (2, y)]$ and \\ $\lambda = 1$, compute the value of $y$. \\ \textbf{Answer Options:} $[-1.61, -0.28, -0.24$ \\ $-0.83]$} & $[-0.24]$ \\
        \hline
        \specialcell{Neural \\ networks I} & \specialcell{\textbf{Question:} If we have a neural network \\ layer with $20$ inputs and $200$ outputs, \\ how many weights (including biases) are \\ needed to describe each connection? \\ \textbf{Answer Options:} $[32000, 80000, 64000,$ \\ $8000]$} & $[64000, 32000, 8000]$ \\
        \hline
    \end{tabular} 
    \caption{Examples of the questions given to the model, the answers generated by the evaluator, and the output answers made by the model in order of the appearance. Note that the final value in output is the correct answer of the question, and the answering stops once the models achieves the correct answer.}
    \label{tab:input-output-examples_mc1}
\end{table}

\begin{table}[!htb]
    \small
    \centering
    \begin{tabular}{|p{3cm}|p{7cm}|l|}
        \hline
        \multicolumn{1}{|c}{Topic} & \multicolumn{1}{|c|}{Question/Multiple Choice Options} & \multicolumn{1}{c|}{Our Model's Answers} \\
        \hline
        \hline
        \specialcell{Neural \\ networks II} & \specialcell{\textbf{Question:} Neuron C is the output \\ neuron which applies a ReLU on its \\ output and neuron A is the input neuron \\ to a neural network. Compute the output \\ of a neural network with the given \\ architecture and inputs. Neuron C takes \\ in the offset value $oC = 1$ with weight \\ $wOC = 3$. Neuron C takes in the output \\ of neuron A with weight $wAC = 1$. \\ Neuron A takes in the input value \\ $x1 = -1$ with weight $w1 = 2$ and offset \\ value $oA = 0.5$ and offset weight \\ $wOC = 3$. \\ \textbf{Answer Options:} $[3, 2.63, 0, 1.5]$} & $[3, 2.63, 1.5]$ \\
        \hline
        \specialcell{Convolutional \\ neural \\ networks} & \specialcell{\textbf{Question:} Using a stride length of $2$, \\ what is the output from applying a filter \\ of length $7$ to an image of length $52$? \\ \textbf{Answer Options:} $[2.2, 25, 22.75, 23]$} & $[23]$ \\
        \hline
        \specialcell{Recurrent \\ neural \\ networks} & \specialcell{\textbf{Question:} An RNN is defined as \\ $s_t = w \times s_{t-1} + x_t$. If $s_0 = 2$, $w = 1$, and \\ $x = [0.25, 0.5]$, what is $s_2$? \\ \textbf{Answer Options:} $[2.75, 0.95, 1.75, 2.63]$} & $[1.75, 0.95, 2.63, 2.75]$ \\
        \hline
        \specialcell{State \\ machines and \\ MDPs} & \specialcell{\textbf{Question:} A state machine is defined by \\ the equations $s_t = f(s_{t-1}, x_t)$ and \\ $y_t = g(s_t)$. Given the conditions $s_0 = 16$, \\ $f(s_{t-1}, x_t) = \max(s_{t-1},x_t)$, and \\ $g(s_t) = 0 * s_t$, compute $y_3$ if the input is \\ $x_t = [8, 7, 9]$. \\ \textbf{Answer Options:} $[9.6, 0, 12.8, 40]$} & $[40, 0]$ \\
        \hline
        \specialcell{Reinforcement \\ learning} & \specialcell{\textbf{Question:} Let $q = 9$. After Q learning, \\ what is $q$ if $a = 0.1$ and $t = 4$? \\ \textbf{Answer Options:} $[3.1, 8.7, 8.5, 8.95]$} & $[8.7, 8.95, 3.1, 8.5]$ \\
        \hline
        Decision trees & \specialcell{\textbf{Question:} What is the entropy of the \\ left side of a region containing $27$ points \\ where the plane has $45$ points in total \\ and $4$ points on the left \\ are positive? \\ \textbf{Answer Options:} $[0.61, 0.299, -0.52, $ \\ $0.297]$} & $[-0.52, 0.61]$ \\
        \hline
    \end{tabular} 
    \caption{Examples of the questions given to the model, the answers generated by the evaluator, and the output answers made by the model in order of the appearance. Note that the final value in output is the correct answer of the question, and the answering stops once the models achieves the correct answer. Thus a single numbers was correct on the first try and longer lists required more attempts.}
    \label{tab:input-output-examples_mc2}
\end{table}

\clearpage
\section{Example Questions with Incorrect Answers}

\begin{table}[!htb]
\centering
\small
\begin{tabular}{|l|c|c|}
\hline
\textbf{Incorrect Questions} & \textbf{Answer} & \textbf{Solution}\\ 
\hline
\hline
1. Given the values for $\theta = 2$ and $\theta_0 = 1$, compute the & $-0.31$ & $-0.05$ \\
NLL loss on the data point $(-2, 0)$. Use log base $e$ of $2.71828$ & & \\
for the log. & & \\
\hline
2. Consider the point $(2, 2)$, the $\theta = 2$ and the& $3.98$ & $4.99$ \\
$\theta_0 = 1$. What is the NLL loss? Use natural log, where the & & \\
base is $2.71828$. & & \\
\hline
3. Given a loss function, $(-2\theta+3)^4$, for gradient descent, & $2.4$ & $0.4$ \\
compute the updated $\theta$ value after one gradient descent step. & & \\
Let $\theta = 2$ and $\eta = 0.05$. & & \\
\hline
4. The optimal $\theta$ value computed by mean squared error is $1$ & $-0.345$ & $2.75$ \\
using the datapoints $[(0, -1), (1, -2), (2, y)]$. If $\lambda = 0.5$, what & & \\
is $y$? & & \\
\hline
5. Compute the value returned from aligning the filter $[1, 4, 1]$ & $21$ & $18$ \\
to the image $[1, 4, 1]$ on top of one another. & & \\
\hline
6. What is the length of the result from applying $F$ to $I$ if $F$ & $34$ & $74$ \\
has length $17$ and $I$ has length $90$? & & \\
\hline
\end{tabular}
\caption{Example questions our model incorrectly answers, its answer, and the solution to the question.}
\label{tab:results_incorrect}
\end{table}

\section{Data Augmentation}

\begin{table}[!htb]
    \centering
    \begin{tabular}{|c|c|}
        \hline
        \textbf{Description} & \textbf{Text}\\
        \hline
        \hline
        Question & Let $s_0 = 0$, $w = 1.5$, and $x = (2, 2, 0)$. Compute $s_3$ if \\
        & $s_t = w * s_{t-1} + x_t$. \\
        \hline
        Template & Let $s_0 = \{a\}$, $w = \{b\}$, and $x = \{c\}$. Compute $s_{|\{c\}|}$ if  \\
        & $s_t = w * s_{t-1} + x_t$. \\
        \hline
        Paraphrased Template & An RNN is defined as $s_{t} = w*s_{t-1} + x_{t}$. If $s_{0}$ is $\{a\}$, $w$ is $\{b\}$, \\
        & and $x$ is $\{c\}$, what is $s_{|\{c\}|}$?\\
        \hline
        Paraphrased Question & An RNN is defined as $s_{t} = w*s_{t-1} + x_{t}$. If $s_{0}$ is $0$, $w$ is $1.5$, \\ & and $x$ is $(2,2,0)$, what is $s_{3}$?\\
        \hline
        Augmentation & An RNN is defined as $s_{t} = w*s_{t-1} + x_{t}$. If $s_{0}$ is $1$, $w$ is $0.5$, \\
        & and $x$ is $(0.25, 0.25)$, what is $s_2$?\\
        \hline
        Expression & $s_{3} = w * (w * (w * s_{0} + x_1) + x_2) + x_3$ \\
        \hline
        Values & $s_{3} = 1.5 * (1.5 * (1.5 * 0 + 2) + 2) + 0$ \\
        \hline
        Answer & 7.5 \\
        \hline
    \end{tabular}
    \caption{An example of converting an original question into a new one containing different values and phrasing. Note that $| \cdot |$ represents taking the length of the input.}
    \label{tab:data_augmentation}
\end{table}

\section{Transformer Hyperparameters}

\begin{table}[!htb]
    \centering
    \small
    \begin{tabular}{|l|c|}
        \hline
        \textbf{Hyperparameters} & \textbf{T5 Transformer}\\
        \hline
        Learning rate & 1e-4 \\
        Batch size & 32 \\
        Epochs & 25 \\
        Number of embeddings & 100 \\
        Number of hidden layers & 512 \\
        Number of layers & 3 \\
        Number of heads & 8 \\
        \hline
    \end{tabular}
    \caption{Hyperparameters of the T5 Transformer.}
    \label{tab:hyperparameters}
\end{table}

\section{Reproducibility}
In the spirit of reproducible research, we make our code available in the supplementary material, and our models and data available \footnote{\href{https://osf.io/eryg7/?view_only=edec7fa83ee74b1eb5cffe25555c4d89}{https://osf.io/eryg7/?view\_only=edec7fa83ee74b1eb5cffe25555c4d89}}.

\end{document}